\documentclass{article}

\usepackage{microtype}
\usepackage{dsfont}
\usepackage{graphicx}
\usepackage{subfigure}
\usepackage{booktabs} 
\RequirePackage{titletoc}

\usepackage{hyperref}

\usepackage[accepted]{icml2024}

\usepackage{amsmath}
\usepackage{amssymb}
\usepackage{mathtools}
\usepackage{amsthm}
\usepackage{amsfonts}
\usepackage{mathrsfs}
\usepackage{multirow}
\usepackage{enumitem}

\usepackage[capitalize,noabbrev]{cleveref}

\usepackage[normalem]{ulem} 

\usepackage[textsize=tiny]{todonotes}

\newcommand{\prob}{\mathrm{P}}
\newcommand{\unif}{\mathrm{U}}
\newcommand{\expec}{\mathbb{E}}
\newcommand{\esp}{\expec}

\newcommand{\sphere}{\mathbb{S}^{d-1}}

\newcommand{\Normal}{\mathcal{N}}

\newcommand{\beq}{\begin{equation}}
\newcommand{\eeq}{\end{equation}}

\def\rset{\mathbb{R}}

\def\nset{\mathbb{N}}

\newcommand{\oh}{o}
\newcommand{\Oh}{\mathcal{O}}

\newcommand\inner[2]{\left \langle #1, #2 \right \rangle}
\newcommand{\Int}{\mathrm{I}}

\definecolor{britishracinggreen}{rgb}{0.0, 0.26, 0.15}

\newcommand{\diff}{\mathrm{d}}

\newcommand{\rbr}[1]{\left({#1}\right)}

\newcommand{\Wass}{\mathrm{W}}
\newcommand{\SW}{\operatorname{SW}}
\newcommand{\slice}{\SW}
\newcommand{\SHCV}{\operatorname{SHCV}}
\newcommand{\Xset}{\mathsf{X}}
\newcommand{\Lip}{\operatorname{Lip}}

\newcommand{\bA}{\mathbf{A}}
\newcommand{\bB}{\mathbf{B}}

\newtheorem{theorem}{Theorem}
\newtheorem{corollary}{Corollary}
\newtheorem{definition}{Definition}
\newtheorem{lemma}{Lemma}

\newtheorem{proposition}{Proposition}

\theoremstyle{definition}
\newtheorem{remark}{Remark}

\newcommand{\Span}{\operatorname{Span}}

\newcommand{\1}{\mathds{1}}

\newcommand{\Tr}{\operatorname{Tr}}

\newcommand{\argmin}{\operatornamewithlimits{\arg\min}}

\newcommand{\Bures}{\mathfrak{B}}
\usepackage{soul}

\definecolor{darkteal}{rgb}{0,0.35,0.35}

\definecolor{mypink1}{rgb}{0.858, 0.188, 0.478}

\definecolor{col_mc}{HTML}{0173B2}
\definecolor{col_cv_low}{HTML}{029E73}
\definecolor{col_cv_up}{HTML}{DE8F05}
\definecolor{col_cvnn}{HTML}{D55E00}
\definecolor{col_shcv}{HTML}{CC78BC}
\definecolor{col_sobol}{HTML}{ECE133}
\definecolor{col_halton}{HTML}{CA9161}

\def\fmn{f^{(p)}_{\mu,\nu}}
\begin{document}

\twocolumn[
\icmltitle{Sliced-Wasserstein Estimation with Spherical Harmonics as Control Variates}



\icmlsetsymbol{equal}{*}

\begin{icmlauthorlist}
\icmlauthor{Rémi Leluc}{cmap}
\icmlauthor{Aymeric Dieuleveut}{cmap}
\icmlauthor{François Portier}{crest}
\icmlauthor{Johan Segers}{isba}
\icmlauthor{Aigerim Zhuman}{isba}
\end{icmlauthorlist}

\icmlaffiliation{cmap}{CMAP, \'Ecole Polytechnique, Institut Polytechnique de Paris, France}
\icmlaffiliation{crest}{CREST, ENSAI, France}
\icmlaffiliation{isba}{ISBA/LIDAM, UCLouvain, Belgium}

\icmlcorrespondingauthor{Rémi Leluc}{remi.leluc@gmail.com}

\icmlkeywords{Optimal Transport, Sliced-Wasserstein, Control Variate, Spherical Harmonics}

\vskip 0.3in
]



\printAffiliationsAndNotice{}  

\begin{abstract}
The Sliced-Wasserstein (SW) distance between probability measures is defined as the average of the Wasserstein distances resulting for the associated one-dimensional projections. As a consequence, the SW distance can be written as an integral with respect to the uniform measure on the sphere and the Monte Carlo framework can be employed for calculating the SW distance. Spherical harmonics are polynomials on the sphere that form an orthonormal basis of the set of square-integrable functions on the sphere. Putting these two facts together, a new Monte Carlo method, hereby referred to as Spherical Harmonics Control Variates (SHCV), is proposed for approximating the SW distance using spherical harmonics as control variates. The resulting approach is shown to have good theoretical properties, e.g., a no-error property for Gaussian measures under a certain form of linear dependency between the variables. Moreover, an improved rate of convergence, compared to Monte Carlo, is established for general measures. The convergence analysis relies on the Lipschitz property associated to the SW integrand. Several numerical experiments demonstrate the superior performance of SHCV against state-of-the-art methods for SW distance computation.
\end{abstract}

\section{Introduction}
\label{sec:intro}

The Sliced-Wasserstein (SW) distance between two probability measures, introduced in \citet{rabin2012wasserstein}, shares similar topological properties with the standard Wasserstein distance \citep{bonnotte2013unidimensional,Bayraktar_eq_W,nadjahi2020statistical}
 while having better properties in terms of computational complexity, especially for measures defined on spaces with relatively large dimension \cite{nadjahi2020statistical}. 
As such, it has emerged as a powerful framework for analyzing and solving a wide range of machine learning problems including 
 generative modeling \cite{deshpande2018generative,deshpande2019max,liutkus2019sliced}, autoencoders \citep{kolouri2018sliced}, Bayesian computation \citep{nadjahi2020approximate}, and image processing \citep{bonneel2015sliced}.

Informally, for $p\geq1$ the $\SW_p$ distance between probability distributions $\mu$ and $\nu$ on $\mathbb{R}^d$ is defined as the integral value, over $\theta$  uniformly distributed on the unit sphere $ \mathbb S^{d-1}$, of the Wasserstein distance $\Wass_p$ between the projections of  distributions $\mu$ and $\nu$ on the direction of $\theta$. 
It takes advantage of the fact that (a) the Wasserstein distance between one-dimensional measures has a simple analytical form, being the $L^p([0,1])$-distance between their quantile functions (approximated using empirical quantiles), and (b) the integral is approximated by Monte Carlo sampling (uniformly on the unit sphere). As a consequence, two types of error affect the accuracy of most algorithms estimating the SW distance: first, the statistical error, due to the approximation of the distributions $\mu$ and $\nu$ by their empirical counterparts, which scales as $m^{-1/2}$, with $m$ the size of the samples observed from $\mu$ and $\nu$ \citep{nietert2022statistical}; 
second, the Monte Carlo integration error, which scales as $n^{-1/2}$, with $n$ the number of projections, as demonstrated in \citet{nadjahi2020statistical} and \citet{nietert2022statistical}. Importantly, these  results establish that the sample complexity to approximate the SW distance does not depend on the dimension. As pointed out in \citet{nadjahi2020statistical}, even though the SW sample complexity is reduced compared to Wasserstein, it requires an additional Monte Carlo step which might offset the previous benefits if the associated variance is large.  

In this paper, we propose and study a new approach to improve SW distance computation by improving the Monte Carlo estimation (step~(b) above). The approach follows from a well-known variance reduction principle introducing \textit{control variates} (functions having known integrals) in the Monte Carlo estimate. Following recent papers \cite{oates2017control,portier2019monte,leluc2021control,south2023regularized}, the use of many control variates can significantly reduce the variance of the Monte Carlo method when the integrand $f$ is well-approximated in some $L^2$-space by the control variates. In \citet{portier2019monte} for instance, the obtained convergence rate is $n^{-1/2} \sigma_{\mathrm{cv}}$, where $\sigma_{\mathrm{cv}}^2 $ is the variance of the residual function resulting from approximating $f$ in the control variate space. By allowing the number of control variates to grow with $n$, the residual standard deviation $\sigma_{\mathrm{cv}}$ can vanish asymptotically, leading to an improvement of the $n^{-1/2}$ rate of the Monte Carlo estimate.

As the integral of interest in the SW distance is over the sphere,  we propose to use  as control variates the well-known \textit{spherical harmonics}, which are homogeneous orthogonal polynomials defined on the sphere. Our approach, referred to as Sliced Wasserstein estimates with Spherical Harmonics as Control Variates (SHCV), leverages the property of spherical harmonics to be a an orthonormal basis of $L_2(\mathbb S^{d-1})$ to build a precise and flexible approximation of the integrand leading to a more accurate Monte Carlo estimate. 

\textbf{Related methods.}
Several methods have been introduced for computing the SW distance and improve upon the standard Monte Carlo estimate.  \citet{nguyen2023control} use a \textit{single} quadratic control variate to reduce the variance. This approach significantly differs from  SHCV, which relies on a potentially large number of control variates. Further, for functions defined on $\rset^d$, \citet{leluc2023speeding} use the \textit{leave-one-out} nearest neighbor regession estimate as a control variate. As soon as the integrand is Lipschitz, the rate of convergence is  $n^{-1/2-1/d}$, which improves upon the $n^{-1/2}$ rate of the Monte Carlo estimate. Even though this technique has not been studied for other  spaces than $\rset^d$ (such as the sphere), good practical performance has been reported in \citet{leluc2023speeding} for approximating the SW distance.

In an attempt to improve the Monte Carlo estimate, Quasi-Monte Carlo (QMC) and Randomized Quasi Monte Carlo (RQMC) methods have been proposed to approximate the SW distance \citep{nguyen2023quasimonte} based on deterministic sequences that are well spread over the integration domain. Finally, we mention the orthogonal approach of \citet{nadjahi2021fast} which is based on Gaussian approximation as the dimension $d$ grows, thus targets large dimension, and only applies for $p=2$.
 
\textbf{Contributions.} The main contributions of this paper are summarized as follows:
\vspace{-0.1cm}
\begin{itemize}[topsep=0pt,leftmargin=*]
\item  We propose a novel variance-reduced $\slice$ estimate based on control variate with spherical harmonics. 

\item We derive some theoretical properties as well as the convergence rate in probability on the SHCV integration error, providing rigorous underpinnings for the proposed methodology. Our analysis reveals that SHCV can achieve faster convergence rates than traditional unbiased estimators of the SW distance.

\item In numerical experiments, our approach outperforms not only existing Monte Carlo approaches based on different control variates but also some recently introduced quasi Monte Carlo (QMC) methods (see related methods described above), without prohibitive additional cost.
\end{itemize}


\textbf{Outline.} \Cref{sec:wasserstein_sliced} introduces the mathematical background of (Sliced-)Wasserstein distances while \Cref{sec:mc_spherical} presents MC integration with control variates. In \Cref{sec:shcv}, we introduce our estimator, based on  spherical harmonics. It is empirically evaluated in \Cref{sec:numerical}. \Cref{sec:theoretical} provides a theoretical analysis and \Cref{sec:conclusion} concludes the paper. The appendices in the supplement contain proofs, algorithms and additional numerical results.

\textbf{Notation.} For a Borel set $\Xset \subseteq \rset^d$, $\mathcal{P}(\Xset)$ is the set of probability measures with support contained in $\Xset$. Let $p \in [1, \infty)$ and $\mathcal{P}_p(\Xset) = \{\mu \in \mathcal{P}(\Xset) : \int_{\Xset} \|x\|^p \, \diff \mu(x)<+\infty \}$ be the set of probability measures on $\Xset$ with finite moment of order $p$. The unit sphere on $\rset^d$ is $\sphere = \{ \theta \in \rset^d: \|\theta\|=1\}$ with surface area $|\sphere|=(2\pi^{d/2})/\Gamma(d/2)$, where $\Gamma$ is the Euler gamma function. $\langle \cdot, \cdot\rangle $ is the Euclidean inner product and $\Normal(\mathrm{m},\mathbf{\Sigma})$ is the $d$-Gaussian distribution with mean $\mathrm{m} \in \rset^d$ and positive definite $d \times d$ covariance matrix $\mathbf{\Sigma}$. The Laplace operator is $\Delta=\partial_1^2+\cdots+\partial_d^2$ where $\partial_i$ is the $i$-th partial derivative.  
\section{Optimal Transport Distances: Wasserstein and Sliced-Wasserstein}
\label{sec:wasserstein_sliced}

\subsection{Wasserstein Distance}
The Wasserstein distance $\Wass_p(\mu,\nu)$ of order $p \in [1,\infty)$ between probability measures $\mu, \nu \in \mathcal{P}_p(\rset^d)$ is defined by
\begin{align*}
\Wass_p^p(\mu,\nu) = \inf_{\pi \in \Pi(\mu,\nu)} \int_{\rset^d \times \rset^d} \|x-y\|^p \, \diff \pi(x,y),
\end{align*}
where $\Pi(\mu,\nu) \subset \mathcal{P}(\rset^d \times \rset^d)$ denotes the set of couplings for $(\mu,\nu)$, i.e., probability measures whose marginals with respect to the first and second variables are $\mu$ and $\nu$, respectively. While the Wasserstein distance enjoys attractive theoretical properties \citep[Chapter 6]{villani2009optimal}, it suffers from a high computational cost. When computing $\Wass_p(\mu_m,\nu_m)$ for discrete distributions $\mu_m$ and $\nu_m$ supported on $m$ points, the worst-case computational complexity scales as $\Oh(m^3 \log m)$ \citep{peyre2019computational}. However, for \textit{univariate} distributions $\mu, \nu \in \mathcal{P}_p(\rset)$, the Wasserstein  distance can be is expressed through the quantile functions $F^{-1}$(inverse c.d.f.) as \citep[see][Section~2]{peyre2019computational}
\[
	\Wass_p^p(\mu,\nu) 
	= \int_{0}^1 |F_{\mu}^{-1}(t)-F_{\nu}^{-1}(t)|^p \, \diff t.
\]
In particular, for discrete measures $\mu_m = m^{-1} \sum_{i=1}^m \delta_{x_i}$ and $\nu_m = m^{-1} \sum_{i=1}^m \delta_{y_i}$ with $x_i, y_i \in \rset$, the Wasserstein distance can be computed by sorting the atoms, yielding
\begin{equation}\label{eq:W_1d}
\Wass_p^p(\mu_m,\nu_m) = \frac{1}{m} \sum_{i=1}^m |x_{(i)
} - y_{(i)}|^p,
\end{equation}
where  $(x_{(i)})_{i=1}^m$ and $(y_{(i)})_{i=1}^m$
are the order statistics of $(x_{i})_{i=1}^m$ and $(y_{i})_{i=1}^m$ respectively. The number of operations induced by the sorting step is $\Oh(m \log m)$. 

\subsection{Sliced-Wasserstein Distances}
The Sliced-Wasserstein distance \citep{rabin2012wasserstein,bonneel2015sliced,kolouri2019generalized} alleviates the high computational of Wasserstein distance, by \textit{slicing} distributions and by taking advantage of the fast computation of the Wasserstein distance between \textit{univariate} distributions (see \cref{eq:W_1d}). 
\begin{definition}[Sliced-Wasserstein distances]
	For $\theta \in \sphere$, let $\theta^\star: \rset^d \to \rset$ denote the linear functional $\theta^\star(x) = \langle \theta, x \rangle$ for $x \in \rset^d$. Let $\prob \in \mathcal{P}(\sphere)$ be a probability distribution on the unit sphere. The $\slice$ distance of order $p \in [1,\infty)$ based on $\prob$ is defined for $\mu, \nu \in \mathcal{P}_p(\rset^d)$ as 
\begin{align}\label{eq:def_sw}
\SW_p^p(\mu,\nu,\prob) 
= \int_{\sphere} \Wass_p^p(\theta_{\sharp}^\star \mu,\theta_{\sharp}^\star \nu) \, \diff \prob(\theta),
\end{align}
where for any measure $\xi \in \mathcal{P}(\rset^d)$,  $\theta_{\sharp}^\star \xi$ is the \textit{push-forward} measure by function $\theta^\star$, i.e.~the distribution of the projection $ \theta^\star(X)$ of a random vector $X\in \rset^d$ with distribution~$\xi$.
\end{definition}
We introduce the integrand $f^{(p)}_{\mu,\nu}:\sphere \to \rset$ defined by
\begin{equation}\label{eq:integrand}
    \forall \theta \in \sphere, \quad  f^{(p)}_{\mu,\nu}(\theta) =  \Wass_p^p(\theta_{\sharp}^\star \mu,\theta_{\sharp}^\star \nu).
\end{equation}
 In practice, the SW distance of \Cref{eq:def_sw} is approximated using a standard Monte Carlo estimate by \textit{(i)} sampling a number $n \geq 1$ of independent random directions $\theta_1,\ldots,\theta_n$ from $\prob$ on $\sphere$ and \textit{(ii)} by averaging the values $(f^{(p)}_{\mu,\nu}(\theta_i))_i$. This is described in \Cref{alg:SWMC}. In the paper, we will often work with the uniform distribution on the sphere, denoted by $\unif$ or $\mathcal{U}(\sphere)$.

Computing the SW distance between discrete measures $\mu_m = m^{-1} \sum_{i=1}^m \delta_{x_i}$ and $\nu_m = m^{-1} \sum_{i=1}^m \delta_{y_i}$ with atoms $x_i, y_i \in \rset^d$ amounts to projecting $(x_{i})_{i=1}^m$ and $(y_{i})_{i=1}^m$ along the $n$ random directions $\theta_1,\ldots,\theta_n$ followed by computing the one-dimensional Wasserstein distances using \Cref{eq:W_1d}. This scheme requires $\Oh(n(dm + m \log m))$ operations which is, in general, faster than computing $\Wass_p^p(\mu_m,\nu_m)$, especially for large $m$.

\begin{remark}[Gaussian case when $p = 2$]
\label{remark:GaussSW}
The Wasserstein distance between two Gaussians $\mu=\Normal(a,\bA)$ and $\nu=\Normal(b,\bB)$ is known in closed form \citep{dowson1982frechet} as
$\Wass_2^2(\mu,\nu) = \|a-b\|_2^2 + \Bures^2(\bA,\bB)$ where $\Bures$ is the Bures distance \citep{bhatia2019bures} on positive definite matrices, $\Bures^2(\bA,\bB) = \Tr(\bA) + \Tr(\bB) - 2\Tr[(\bA^{1/2} \bB \bA^{1/2})^{1/2}]$. 
The Sliced-Wasserstein distance between $\mu$ and $\nu$ relies on the projected distributions $\theta^\star_{\sharp}\mu \sim \Normal(\theta^\top a,\theta^\top \bA \theta)$ and $\theta^\star_{\sharp}\nu \sim \Normal(\theta^\top b,\theta^\top \bB \theta)$ via the integration over $\theta \in \sphere$ of 
$\Wass_2^2(\theta_{\sharp}^\star \mu,\theta_{\sharp}^\star \nu) = |\theta^\top(a-b)|^2 + \bigl(\sqrt{\theta^\top \bA \theta}- \sqrt{\theta^\top \bB \theta}\bigr)^2$.
\end{remark}

\begin{algorithm}[t]
\caption{\textcolor{col_mc}{Sliced Wasserstein Monte Carlo}}\label{alg:SWMC}
\begin{algorithmic}[1]
\REQUIRE $\mu, \nu \in \mathcal{P}_p(\rset^d)$, number of random projections $n$, probability $\prob \in \mathcal{P}(\mathbb{S}^{d-1})$
\STATE Sample random projections $\theta_1,\ldots,\theta_n \sim \prob$
\STATE Compute $f_n = (\fmn(\theta_i))_{i=1}^{n}$
\STATE Return average $\textcolor{col_mc}{\text{MC}}_{n} = (\1_n/n)^\top f_n \phantom{(\fmn(\theta_i))_{i=1}^{n}}$
\end{algorithmic}
\end{algorithm}

\section{Monte Carlo with Control Variates}
\label{sec:mc_spherical}

This section presents the mathematical framework of Monte Carlo integration with control variates.

\textbf{Monte Carlo integration.} Consider a square-integrable, real-valued function $f \in L_2(\prob)$ on a probability space $(\Theta,\mathcal{F},\prob)$ of which we would like to compute the integral
\begin{equation}\label{eq:problem}
    \Int(f) 
    = \int_{\Theta} f(\theta) \, \diff\prob(\theta).
\end{equation}
When $\Int(f)$ does not admit a closed form or when only calls from $f$ are available, one may rely on Monte Carlo estimates. Let $\theta_1,\ldots,\theta_n \sim \prob$ be an independent and identically distributed (i.i.d.) random sample from $\prob$. The naive Monte Carlo estimate $\Int_n(f)$ of $\Int(f)$ is given by the empirical mean $$\Int_n(f) = \frac{1}{n} \sum_{i=1}^n f(\theta_i).$$
The Monte Carlo estimate is unbiased and has variance equal to $\sigma^2(f)/n$ where $\sigma^2(f) = \Int[\{f-\Int(f)\}^2]$. Increasing the sample size $n$ reduces the variance but at the cost of an increased computation time. Instead, it is possible to find another estimate with smaller variance.

\textbf{Control Variates and OLSMC.} 
A classical way to reduce the variance of the Monte Carlo estimate consists in incorporating knowledge about some specially chosen functions. Control variates are functions $\varphi_1,\ldots,\varphi_s: \Theta \to \rset$ with known expectations; without loss of generality, we assume that $\Int(\varphi_j)=0$ for $j=1,\ldots,s$.  Let $\varphi = (\varphi_1,\ldots,\varphi_s)^\top$ denote the $\rset^s$-valued function with $s$ control variates as elements. Since all control variates are centered, adding any linear combination of the form $\beta^\top \varphi$ with $\beta \in \rset^s$ gives $\Int(f - \beta^\top \varphi) = \Int(f)$. Thus, given an i.i.d.\ sample $\theta_1,\ldots,\theta_n$ from $\prob$, any coefficient vector $\beta \in \rset^s$ yields an unbiased estimate of $ I(f)$, given by
\begin{equation}\label{eq:class_cv}
    \Int_n(f,\beta) = \frac{1}{n} \sum_{i=1}^n \left(f(\theta_i) - \beta^\top \varphi(\theta_i) \right),
\end{equation}
with variance equal to $\sigma^2(f-\beta^\top \varphi)/n$. 
\begin{figure}[t]
    \centering
    \includegraphics[scale=0.4]{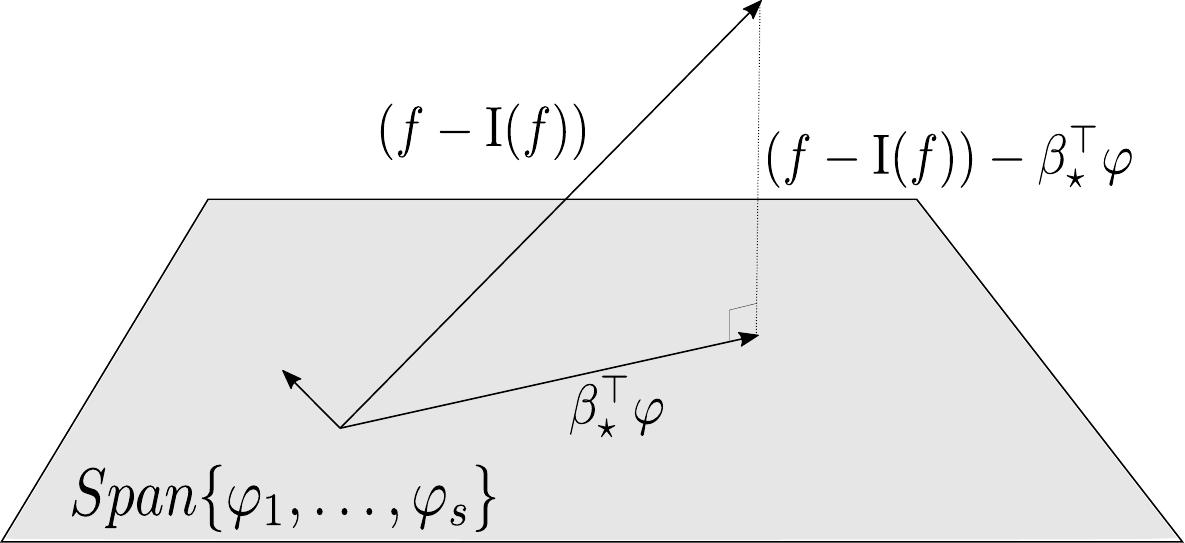}
    \caption{Visualization of the $L_2$-projection of the integrand $f$ onto the linear space $\Span\{\varphi_1,\ldots,\varphi_s\}$ of control variates.}
    \label{fig:L2_proj}
\end{figure}
The optimal coefficient $\beta_\star$ is defined as the one minimizing the variance,
\begin{align*}
    \beta_\star \in \argmin_{\beta \in \rset^s} \Int[\{f - \Int(f)- \beta^\top \varphi\}^2].
\end{align*}
By the Hilbert projection theorem, $\beta_{\star}^\top \varphi$ is the $L_2(\prob)$-projection of $f - \Int(f)$ on the linear space of control variates $\Span\{\varphi_1,\ldots,\varphi_s\}$, as illustrated in \Cref{fig:L2_proj}. 

Different approaches have been proposed to estimate $\beta_{\star}$, each of which, when used in \Cref{eq:class_cv}, produces an enhanced Monte Carlo estimate with reduced variability \citep{glynn2002some}. Following \citet[Section~8.9]{mcbook}, or more recently \citet{portier2019monte} and \citet{leluc2021control}, we frame the problem as an ordinary least-squares minimization problem. The integral $\Int(f)$ appears as the intercept of the linear regression model with explanatory variables $\varphi_1,\ldots,\varphi_s$ and target response $f$, i.e.,
\begin{align*}
    (\Int(f),\beta_\star(f)) \in \argmin\limits_{(\alpha,\beta) \in \rset \times \rset^s} \Int[(f - \alpha- \beta^\top \varphi)^2]. 
\end{align*}
In practice, these quantities are estimated using the directions $\theta_1,\ldots,\theta_n$ sampled from $\prob$. The Ordinary Least Squares Monte Carlo (OLSMC) estimate $\Int_n^{\mathrm{ols}}(f)$ is\footnote{It is well-defined if and only if $\1_n \notin \Span(\Phi)$. If ill-defined, one can reduce the number of control variates such that $s<n$.} 
\begin{equation}\label{eq:olsmc}
    (\Int_n^{\mathrm{ols}}(f),\beta_n(f)) \in \argmin_{(\alpha,\beta) \in \rset \times \rset^s} \|f_n - \alpha \1_n - \Phi \beta\|_2^2, 
\end{equation}
where $f_n = (f(\theta_1),\ldots,f(\theta_n))^\top \in \rset^n$ is the vector of evaluations of $f$, $\1_n 
 = (1,\ldots,1)^\top \in \rset^n$ and $\Phi \in \rset^{n \times s}$ is the random matrix of control variates $\Phi = (\varphi(\theta_i)^\top)_{i=1}^n$.

\begin{remark}[Complexity]
\label{rem:comp_ols}
The Monte Carlo estimate requires $\Oh(n \omega_f)$ computations where $\omega_f$ is the cost of evaluating the integrand $f$. The OLSMC estimate has a complexity of $\Oh(n \omega_f + \omega(\Phi))$ where $\omega(\Phi)$ is the complexity due to the additional step of fitting the optimal control variates. Typically, $\omega(\Phi)=\Oh(ns^2 + s^3)$ where $\Oh(ns^2)$ and $\Oh(s^3)$ operations are needed to compute and invert $\Phi^\top \Phi$, though it can be reduced by using approximate solution via optimization algorithms such as (stochastic) gradient descent algorithms \citep{zhang2004solving}. When $f$ is heavy to compute---which is the case for $f^{(p)}_{\mu_m,\nu_m}$ with discrete measures supported on a large number $m$ of atoms---the extra term $\omega(\Phi)$ becomes negligible compared to $\Oh(n \omega_f)$.
\end{remark}
\begin{remark}[Regularization] 
As the number $s$ of control variates may increase quickly with the dimension, \citet{leluc2021control} rely on the LASSO procedure \citep{tibshirani1996regression} in place of \Cref{eq:olsmc} to achieve estimation and variable selection at the same time in the regime $n \ll s$.
\end{remark}

\section{Sliced-Wasserstein Estimate with Spherical Harmonics as Control Variates}
\label{sec:shcv}

\subsection{Spherical Harmonics}

 Originating from the solutions to the Laplace equation on the unit sphere, spherical harmonics \citep{atkinsonhan,dai2013approximation} constitute a mathematical framework with applications in computer graphics \citep{ramamoorthi2001efficient,basri2003lambertian,green2003spherical} and machine learning \citep{s.2018spherical,dutordoir2020sparse}, particularly in tasks involving spherical data representation.

\begin{definition}[Polynomial spaces]
    Let $\mathscr{P}_{\ell}^d$ be the space of homogeneous polynomials of degree $\ell \geq 0$ on $\rset^d$, i.e., $\mathscr{P}_{\ell}^d = \Span\{x_1^{a_1} \cdots x_d^{a_d} \mid a_k \in \nset, \sum_{k=1}^d a_k = \ell\}$. Let $\mathscr{H}_{\ell}^d$ be the space of real harmonic polynomials, homogeneous of degree $\ell$ on~$\rset^d$, i.e., $\mathscr{H}_{\ell}^d = \{ Q \in \mathscr{P}_{\ell}^d \mid \Delta Q = 0\}$.
\end{definition}

\begin{definition}[Spherical harmonics]
\label{def:nb_har}
Spherical harmonics of degree (or level) $\ell \geq 0$ are defined as the restriction of elements in $\mathscr{H}_{\ell}^d$ to the unit sphere $\sphere$, i.e., restrictions to the sphere of harmonic homogeneous polynomials with $d$ variables of degree $\ell$.
\end{definition}
 Let $N_{\ell}^d$ denote the number of linearly independent spherical harmonics of degree $\ell \ge 0$ in dimension $d$, that is $N_{\ell}^d = \dim \mathscr{H}_{\ell}^d = \dim \mathscr{P}_{\ell}^d - \dim \mathscr{P}_{\ell-2}^d$
\citep[Corollary 1.1.4]{dai2013approximation} where it is agreed that $\dim \mathscr{P}_{\ell-2}^d=0$ for $\ell=0,1$. We have
\begin{align*}
    N_{\ell}^d = \frac{(2\ell + d - 2)(\ell + d - 3)!}{\ell!(d-2)!}\cdot
\end{align*}
Note that, when restricted to the sphere, all polynomials are linear combinations of spherical harmonics. Moreover, spherical harmonics of different degrees are orthogonal with respect to $ \langle f,g \rangle  =  \int_{\sphere} f(\theta) g(\theta) \, \diff \unif(\theta)$, where $\unif$ is the uniform distribution on $\sphere$ \citep[Theorem 1.1.2]{dai2013approximation}. In practice, orthonormal bases can be constructed by applying the Gram–Schmidt process and an explicit basis of spherical harmonics can be written in terms of the Gegenbauer polynomials in spherical coordinates \citep[Theorem 1.5.1]{dai2013approximation}. In the following, let $\varphi_{\ell,k}$ with $\ell \ge 0$ and $k=1, \ldots,N_{\ell}^d$ denote a set of spherical harmonics forming an orthonormal basis of $\mathscr{H}_{\ell}^d$. 

\begin{proposition}[Hilbert basis]\label{prop:approx}
    The spherical harmonics $\{\varphi_{\ell,k}, 0\le \ell, 1\le k \le N_\ell^{d}\}$ form an orthonormal eigenbasis of the Hilbert space $L_2(\sphere, \unif)$, so that for every $f \in L_2(\sphere)$ we have
    \[
        f = \sum_{\ell=0}^{\infty} \sum_{k=1}^{N_{\ell}^d} \hat f_{\ell,k} \varphi_{\ell,k}
        \quad \text{where} \quad
        \hat f_{\ell,k} = \int f \varphi_{\ell,k} \, \diff \unif.
    \]
\end{proposition}

Spherical harmonics thus constitute an eigenbasis of centered functions, and can therefore be used as control variates with respect to the uniform distribution on the unit sphere.
For $\ell \geq 1$ and $k=1,\ldots,N_{\ell}^d$, we have
\begin{equation}
\label{eq:Iphilk0}	
	\Int(\varphi_{\ell,k})
	= \int_{\sphere} \varphi_{\ell,k}(\theta) \, \diff\unif(\theta)
	= 0.
\end{equation}
Orthonormality ensures that the empirical covariance matrix is well-conditioned when solving the OLS problem \eqref{eq:olsmc}.

\begin{remark}[Examples] \label{rem:class_ex} In most applications, spherical harmonics in two and three variables are used. For $d=2$, we have $\dim \mathscr{H}_{\ell}^2=2$ for all $\ell \ge 1$ and an orthogonal basis is given by the real and imaginary parts of $(x+\mathrm{i}y)^{\ell}$. In polar coordinates $(x,y)=(r \cos(t),r \sin(t)) \in \rset^2$, this basis is $\varphi_{\ell,1}(x,y)=r^\ell \cos(\ell t)$ and $\varphi_{\ell,2}(x,y)=r^\ell \sin(\ell t)$. By restriction on the circle $\mathbb{S}^1$, spherical harmonics yield the classical Fourier expansions in cosine and sine functions and are commonly referred to as \textit{circular harmonics}. For $d=3$, we have $\dim \mathscr{H}_{\ell}^3=2\ell + 1$ and a classical orthogonal basis in spherical coordinates is written in terms of the associated Legendre polynomials (see \cref{subsec:spherical_small_dim}).
\end{remark}

\subsection{Spherical Harmonics Control Variates Estimate}
With the idea of performing variance reduction when computing $\slice$ distances, we incorporate spherical harmonics as natural control variates  on $\sphere$ into the standard Monte Carlo estimate. We propose a novel procedure called \textit{Spherical Harmonics Control Variates} (SHCV) estimate using the ordinary least squares formulation of \Cref{eq:olsmc}. 

For $p\geq 1$, consider $\mu, \nu \in \mathcal{P}_p(\rset^d)$ and recall the integrand $f^{(p)}_{\mu,\nu}: \theta \mapsto  \mathrm{W}_p^p(\theta_{\sharp}^\star \mu,\theta_{\sharp}^\star \nu)$ so that $\SW_p^p(\mu,\nu) = \Int(f_{\mu,\nu}^{(p)})$. Interestingly, this integrand is even so that for odd functions $\varphi$, the inner product vanishes, $\langle\fmn,\varphi\rangle = 0$. The spherical harmonics expansion of $\fmn$ (see \Cref{prop:approx}) is only made of spherical harmonics with even degree. Thus, when finding the optimal control variates among spherical harmonics, we can omit the ones of odd degree as these are uncorrelated with our integrand.

\begin{lemma}[Number of control variates]
\label{lem:counting}
In dimension $d$, the dimension of the linear space generated by the spherical harmonics of even degree up to maximal degree $2L$ is
\[
	s_{L,d} = \sum_{\ell =1}^{L} N^d_{2\ell}
	= \binom{2L+d-1}{d-1} - 1.
\]
\end{lemma}
For $d = 2$, the formula specializes to $s_{L,2}=2L$ which is the number of circular harmonics of degree $2, 4, \ldots, 2L$; For $d = 3$, the formula becomes $s_{L,3}=(2L+3)L$, which is indeed the sum of $(2\ell+1)$ for $\ell = 2, 4, \ldots, 2L$ (\Cref{rem:class_ex}).

\begin{definition}[SHCV]\label{def:shcv}
For $\mu, \nu \in \mathcal{P}_p(\rset^d)$,
the SHCV estimate of maximum degree $2L$ based on projections $\theta_1,\ldots,\theta_n \in \sphere$ and control variates $\varphi =(\varphi_{j})_{j =  1} ^{s_{L,d}}$ is defined as the intercept of the OLS problem \eqref{eq:olsmc} with the integrand $f_{\mu,\nu}^{(p)}$ as response and all spherical harmonics of even degree from $2$ up to $2L$ as $n \times s_{L,d}$ covariate matrix $\Phi = (\varphi (\theta_i)^\top)_{i =  1}  ^n$, i.e.,
\begin{equation}\label{eq:shcv}
\SHCV_{n,L}^p({\mu,\nu}) = \Int_{n}^{\mathrm{ols}}(f_{\mu,\nu}^{(p)}).
\end{equation}
\end{definition}

The procedure is described in  \Cref{algo:shcv}. In the algorithm, the number of spherical harmonics is specified in advance. A selection based on a lasso procedure with cross-validation for the penalization parameter would be possible too, but is not pursued here for the sake of computing efficiency. 

\begin{algorithm}[t]
\caption{\textcolor{col_shcv}{Spherical Harmonics Control Variate Estimate}}
\begin{algorithmic}[1]
\REQUIRE $\mu, \nu \in \mathcal{P}_p(\rset^d)$, number of random projections $n$, $\prob \in \mathcal{P}(\mathbb{S}^{d-1})$, spherical harmonics $\varphi = (\varphi_{j})_{j =  1} ^{s}$
\STATE Sample random projections $\theta_1,\ldots,\theta_n \sim \prob$
\STATE Compute $f_n = (\fmn(\theta_i))_{i=1}^{n}$, $\Phi = (\varphi (\theta_i)^\top)_{i =  1}  ^n$
\STATE Solve $(\Int_n^{\mathrm{ols}},\beta_n) \in \arg\min_{\alpha,\beta} \|f_n - \alpha \1_n - \Phi \beta \|_2^2$ 
\STATE Return $\textcolor{col_shcv}{\mathrm{SHCV}}_{n} = \Int_n^{\mathrm{ols}}$ 
\end{algorithmic}
\label{algo:shcv}
\end{algorithm}

\begin{remark}[Multiple integrals]\label{rem:multiple}  Computational benefits occur when there are multiple integrands, since the SHCV estimate can be represented as a linear rule $w^\top f_n$ \citep{portier2019monte,leluc2021control}, where the weight vector $w \in \rset^n$ does not depend on the integrands. If $K$ integrals need to be evaluated, the computation time of SHCV becomes $\Oh(K n \omega_f + \omega(\Phi))$ compared to $\Oh(K n \omega_f)$ for standard Monte Carlo, and the additional cost $\omega(\Phi)$ becomes negligible for large $K$. This is particularly relevant for $\slice$-based kernels involving many pairwise computations $\slice_2^2(\mu_i,\nu_j)$; see \Cref{alg:MC_kernel,alg:SHCV_Kernel} in \Cref{subsec:kernel_svm}.
\end{remark}

\begin{remark}[Quadratic control variates]
 Any symmetric $d \times d$ matrix $M$ yields a quadratic control variate $\varphi_M(\theta) = \theta^\top M \theta - \Tr(M)/d$ as $\Int(\varphi_M) =0 $. This is actually the approach used in \citet{nguyen2023control} where the authors propose two different choices for $M$ leading to two different estimators (see \Cref{app:algos}). Note that, whenever $2L\geq 2$, these quadratic functions are included in our spherical harmonics space, making our method more general with potentially greater variance reduction (see \Cref{sec:theoretical}). 
\end{remark}

\section{Numerical Experiments}
\label{sec:numerical}

To asses the finite-sample performance of the SHCV estimate, we first present some synthetic data examples involving the computation of $\slice_2^2(\mu,\nu)$ between multivariate Gaussian distributions. We consider different dimensions $d \in \{3;5;6;10;20\}$ and a varying number of random projections $n \in [10^2;10^4]$. Then we focus on $\slice_2^2(\mu_m,\nu_m)$ between 3D point clouds from the \textsf{ShapeNetCore} dataset of \citet{chang2015shapenet}. In both cases, the ground truth of the Sliced-Wasserstein distance is estimated with a very large number $n=10^8$ of projections. The different results report the mean squared error (MSE) $\expec[|{{\slice}_{2,n}^{2}}(\mu,\nu)-\slice_2^2(\mu,\nu)|^2]$ for the different estimates ${{\slice}_{2,n}^{2}}(\mu,\nu)$ where the expectation is computed as an average over $100$ independent runs. Finally, in the spirit of \citet{kolouri2016sliced} and \citet{meunier2022distribution}, we implement a Kernel Support Vector Machine (SVM) classifier for image classification. For ease of reproducibility, numerical details are in the appendix and the code is available \href{https://github.com/RemiLELUC/SHCV}{here}. In particular, we rely on the efficient implementation of \citet{dutordoir2020sparse} to build the spherical harmonics. The degrees of the spherical harmonics are reported in \Cref{tab:cv_number} in \Cref{sec:cv_number}.

\paragraph{Methods in competition.} In the experiments, the $\slice$ distance is computed through different estimates. Along with the standard Monte Carlo estimate and the proposed SHCV estimate, we consider three other baselines based on control variates: the two estimates of \citet{nguyen2023control} based on lower and upper bounds of a Gaussian approximation and the \textit{control neighbors} estimate of \citet{leluc2023speeding} based on nearest neighbors estimates acting as control variates. For the sake of completeness, we also include a comparison with (Randomized) Quasi Monte Carlo procedures as in \citet{nguyen2023quasimonte} where deterministic QMC sets from $[0,1]^d$ are mapped to the unit sphere and randomly rotated. The methods in competition are: the standard Monte Carlo estimate (\textcolor{col_mc}{MC}), the lower-CV (\textcolor{col_cv_low}{$\text{CV}_{low}$}) and upper-CV (\textcolor{col_cv_up}{$\text{CV}_{up}$}) estimates, the control neighbors estimate (\textcolor{col_cvnn}{CVNN}), the proposed estimate with spherical harmonics as control variates (\textcolor{col_shcv}{SHCV}), the Quasi Monte Carlo (\textcolor{col_sobol}{QMC}) and Random Quasi Monte Carlo estimates (\textcolor{col_halton}{RQMC}).

 \begin{figure}[t]
  \centering
  \subfigure[$d=3$]{
  \includegraphics[scale=0.41]{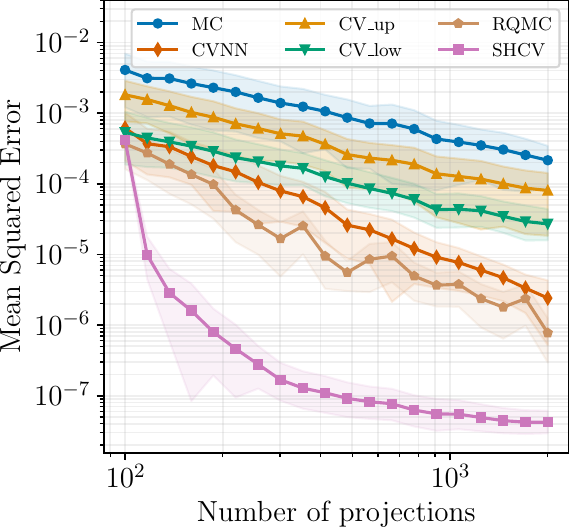}\label{fig:d3}}
  \subfigure[$d=6$]{
  \includegraphics[scale=0.41]{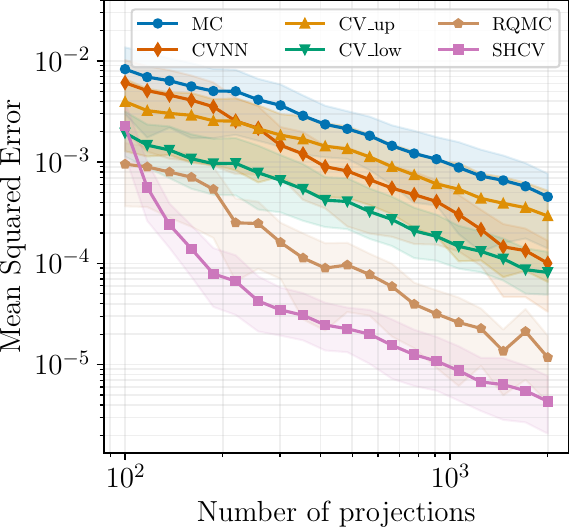}\label{fig:d6}}
\caption{MSE curves for sampled Gaussian distributions in dimension $d \in \{3;6\}$.}
    \label{fig:compare_gaussian_proj}
\end{figure}

\subsection{Synthetic Data: Multivariate Gaussians}

We compute $\slice_2^2(\mu_m,\nu_m)$ between empirical distributions $\mu_m=m^{-1}\sum_{i=1}^m \delta_{x_i}$ and $\nu_m=m^{-1}\sum_{j=1}^m \delta_{y_j}$ where $x_1,\ldots,x_m \sim \mu=\Normal(a,\mathbf{A})$, $y_1,\ldots,y_m \sim \nu=\Normal(b,\mathbf{B})$ with $m=1000$ samples. The parameters of the Gaussians are $a,b \sim \Normal(\1_d,I_d)$ and covariance matrices $\mathbf{A}=\Sigma_a^{}\Sigma_a^\top$ and $\mathbf{B}=\Sigma_b^{}\Sigma_b^\top$ where all entries of $\Sigma_a, \Sigma_b$ are drawn according to $\Normal(0,1)$. In  \Cref{subsec:exact_app}, we also consider the \textit{exact case} where we directly evaluate the smooth integrand  $f_{\mu,\nu}^{(2)}$ given in \Cref{remark:GaussSW}. In that case, as the original lower-CV (\textcolor{col_cv_low}{$\text{CV}_{low}$}) and upper-CV (\textcolor{col_cv_up}{$\text{CV}_{up}$}) estimates are written to take discrete measures as inputs, we adapt them with the true means and covariance matrices.

\Cref{fig:compare_gaussian_proj} reports the MSE of the different methods in dimension $d\in \{3;6\}$ with respect to the number of projections, while \Cref{fig:synthetic_sampled_appendix} in  \Cref{subsec:sampled} does so with respect to computing time. \Cref{tab:res_gaus_dim_exact} reports the MSE and computing times (in ms) in higher dimension $d\in \{5;10;20\}$. \Cref{fig:synthetic_exact_appendix} and  \Cref{tab:res_gaus_dim_exact_appendix} in  \Cref{subsec:exact_app} deal with the \textit{exact case} accordingly. In each case, the SHCV estimate gives the best MSE performance with respect to both the number of projections and computing time, beating other baselines by a factor up to $100$.

\begin{table}[t]
  \centering
  \resizebox{0.49\textwidth}{!}{
  \begin{tabular}{lcccccc}

    \toprule
     \multirow{2}{2em}{Method} & \multicolumn{2}{c}{$d=5$}  & \multicolumn{2}{c}{$d=10$} & \multicolumn{2}{c}{$d=20$} \\
    \cmidrule(l){2-3}\cmidrule(l){4-5}\cmidrule(l){6-7}
    & MSE & Time & MSE & Time & MSE & Time \\
    \midrule
     \textcolor{col_mc}{MC} & $1.45$e-$3$ &  $81.1\pm3.5$ & $9.45$e-$4$ & $80.7\pm4.4$ & $1.47$e-$3$ & $81.1\pm1.8$  \\
    \textcolor{col_cv_low}{$\text{CV}_{\text{low}}$} & $2.67$e-$4$ & $79.7\pm1.1$ & $3.45$e-$4$ & $80.1\pm1.4$ & $3.82$e-$4$ & $80.0\pm1.0$ \\
    \textcolor{col_cv_up}{$\text{CV}_{\text{up}}$}&  $8.44$e-$4$ & $83.0\pm1.2$  & $7.51$e-$4$ & $83.0\pm1.7$ & $1.09$e-$3$ & $83.1\pm1.5$   \\
    \textcolor{col_cvnn}{CVNN}  & $4.29$e-$4$ & $110\phantom{.}\pm2.2$ & $1.12$e-$3$ & $122\phantom{.}\pm1.6$ & $2.14$e-$3$ & $127\phantom{.}\pm1.4$ \\
    \textcolor{col_sobol}{QMC}  & $2.91$e-$4$ & $100\phantom{.}\pm1.2$ & $2.37$e-$4$ & $113\phantom{.}\pm1.4$ & $6.60$e-$4$ & $129\phantom{.}\pm1.4$ \\
    \textcolor{col_halton}{RQMC}  & $5.80$e-$5$ & $96.3\pm2.2$ & $2.75$e-$4$ & $113\phantom{.}\pm1.2$ & $1.17$e-$3$ & $130\phantom{.}\pm1.0$  \\
    \textcolor{col_shcv}{SHCV} & $\mathbf{2.68}$e-$\mathbf{6}$ & $89.0\pm6.3$ & $\mathbf{1.93}$e-$\mathbf{4}$ &  $89.0\pm4.5$ & $\mathbf{2.95}$e-$\mathbf{4}$ & $88.1\pm2.8$  \\
    \bottomrule
  \end{tabular}}
  \caption{MSE and computing time (ms) for Gaussian distributions in dimension $d \in \{5;10;20\}$ based on $n=500$ projections.}
  \label{tab:res_gaus_dim_exact}
\end{table}

\subsection{Empirical Distributions of 3D Point Clouds}

 \begin{figure*}[t]
  \centering
  \subfigure[\textsc{plane/lamp}]{
  \includegraphics[scale=0.555]{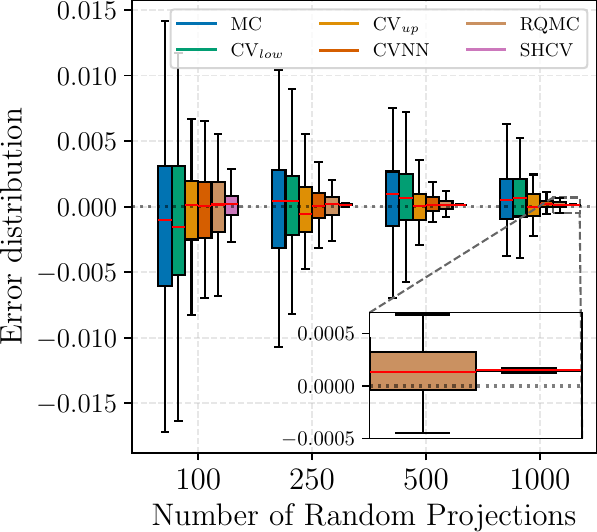}\label{fig:box1}}
   \subfigure[\textsc{lamp/bed}]{
  \includegraphics[scale=0.555]{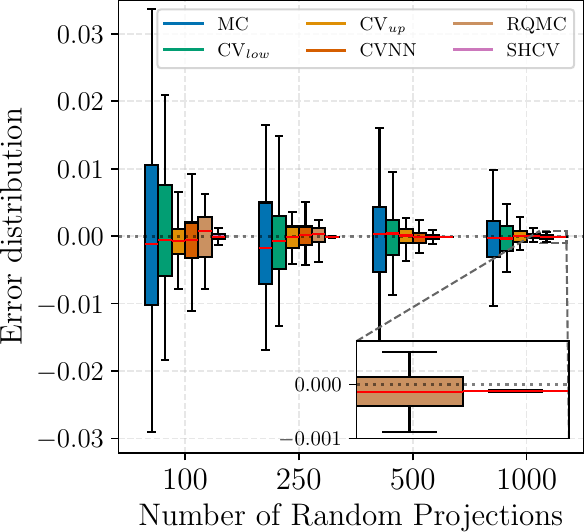}\label{fig:box2}}
  \subfigure[\textsc{plane/bed}]{
  \includegraphics[scale=0.555]{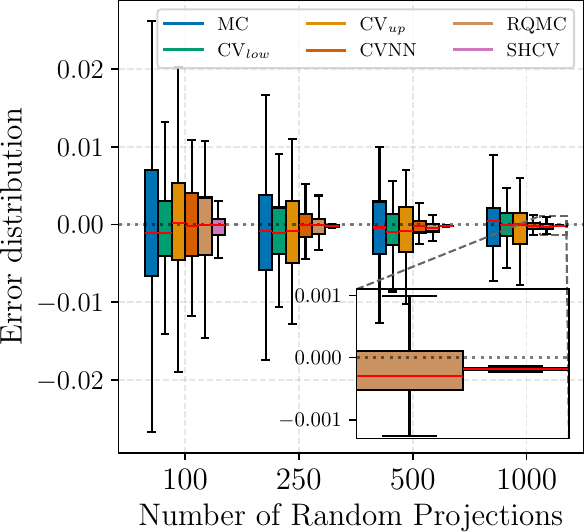}\label{fig:box3}}
\vspace{-0.2cm}\caption{Boxplots of the error distribution $\widehat{\slice}_n(\mu_m,\nu_m)-\slice(\mu_m,\nu_m)$ for different SW estimates based on $n$ random projections with $n \in \{100;250;500;1000\}$ obtained over $100$ independent runs.}
    \label{fig:box}
\end{figure*}

Similarly to \citet{nguyen2023control}, we use three point clouds taken at random from the dataset \textsf{ShapeNetCore} \citep{chang2015shapenet} corresponding to the objects \textsf{plane}, \textsf{lamp}, and \textsf{bed}, each composed of $m=2048$ points in $\rset^3$. Each point cloud $(x_1,\ldots,x_m)$ is encoded as a discrete measure $\mu_m = m^{-1}\sum_{i=1}^m \delta_{x_i}$ with equal masses over the points in the cloud and the goal is to compute $\slice(\mu_m,\nu_m)$ of order $p=2$ between empirical distributions. \Cref{fig:box} reports the boxplots of the error distribution 
$\widehat{\slice}_n(\mu_m,\nu_m)-\slice(\mu_m,\nu_m)$ for the different SW estimates based on $n$ random projections with $n \in \{100;250;500;1000\}$ obtained over $100$ independent runs. Once again, our SHCV estimate provides huge gains in terms of variance reduction and outperforms other baselines by a factor $100$; see \cref{tab:mse_clouds} in \Cref{subsec:app_clouds}.

\subsection{Sliced-Wasserstein and Kernel SVM}

$\slice$ distances are useful in the context of distribution regression \citep{szabo2016learning} where the goal is to learn a real-valued function defined on the space of probability distributions from a sequence of observations. \citet{kolouri2016sliced} first introduced a $\slice$-based kernel for absolutely continuous distributions, which was later on extended to empirical distributions by \citet{meunier2022distribution}. For any $\gamma>0$, both $e^{-\gamma \SW_2^2(\cdot,\cdot)}$ and $e^{-\gamma \SW_1(\cdot,\cdot)}$ are valid kernels on probability measures. Using the $\slice$ kernel of \citet{kolouri2016sliced} defined by $k(\mu_i,\mu_j) = \exp(-\gamma \slice_2^2(\mu_i,\mu_j))$ with $\gamma >0$, we consider the image classification task on the \textsf{digits} dataset of \citet{digits_dataset}. This dataset contains $N=1\,797$ images of size $8 \times 8$ and the goal is to predict the label $z \in \{0,1,\ldots,9\}$ of the handwritten digit. Similarly to \citet{meunier2022distribution}, we convert each image to a histogram so that the $\slice$ kernels have probability distributions as inputs. For image $i \in [N]$, let $(x_k^i, y_k^i)_{k=1}^{n_i}$ denote the coordinates of the $n_i$ active pixels and $(\psi_k^i)_{k=1}^{n_i}$ the associated pixel intensities. The positions are renormalized to the grid $[0,1]^2$ and we consider the weighted histogram $\mu_i = \sum_{k=1}^{n_i} \alpha_k^i \delta_{(x_k^i,y_k^i)}$ where $\alpha_k^i = \psi_k^i/\sum_{l=1}^{n_i} \psi_l^i$ are the normalized intensities. We use a train/test split of size $80/20$ giving $N_{\text{train}}=1\,437$ training and $N_{\text{test}}=360$ testing images, respectively, where each set is balanced. 

We implement a Kernel SVM classifier where the $\slice_2$ distance involved in the kernel expression is either computed with the standard Monte Carlo estimate (\textcolor{col_mc}{MC}) or with the SHCV estimate (\textcolor{col_shcv}{SHCV}). The number of random projections $n$ evolves in $\{25;50;75;100\}$. To compare different scenarios, the bandwidth parameter $\gamma>0$ of the Gaussian kernel is in $\{0.01;1\}$ and the regularization parameter $C$ of the SVM is optimized with a cross-validation strategy.  \Cref{tab:kernel_svm} reports the means and standard deviations of the test accuracies of the different methods where the statistics are obtained over $10$ independent runs.
\begin{table}[h!]
  \centering
  \resizebox{0.48\textwidth}{!}{
  \begin{tabular}{ccccc}
    \toprule
    \multirow{2}{*}{$n$} & \multicolumn{2}{c}{$\gamma=0.01$} & \multicolumn{2}{c}{$\gamma=1$}  \\
    \cmidrule(l){2-3}\cmidrule(l){4-5}
    & \textcolor{col_mc}{MC} & \textcolor{col_shcv}{SHCV} & \textcolor{col_mc}{MC} & \textcolor{col_shcv}{SHCV} \\
    \midrule
    $25$ & $86.0\pm0.6$ & $\mathbf{87.3}\pm0.3$ & $92.7\pm0.6$ & $\mathbf{93.3}\pm0.6$   \\
    $50$ & $86.7\pm0.5$ & $\mathbf{87.6}\pm0.3$ & $93.2\pm0.4$ & $\mathbf{94.0}\pm0.3$  \\
    $75$ & $86.8\pm0.2$ & $\mathbf{87.8}\pm0.2$ & $93.3\pm0.4$ & $\mathbf{94.3}\pm0.2$  \\
    $100$ & $86.9\pm0.3$ & $\mathbf{87.7}\pm0.1$ & $93.3\pm0.3$ & $\mathbf{94.3}\pm0.2$  \\
    \bottomrule
  \end{tabular}}
  \caption{Kernel SVM: test accuracies with means and standard deviations. The statistics are obtained over $10$ independent runs. \vspace{-0.3cm}}
  \label{tab:kernel_svm}
\end{table}

\section{Theoretical Analysis}
\label{sec:theoretical}

This section gathers several theoretical properties of the novel SHCV estimate. We first state some elementary properties of the underlying integration rule with respect to a general distribution $\prob$ on the sphere and then provide a convergence rate in probability of the integration error in case of the uniform distribution $\unif$ on the sphere.

\subsection{Elementary Properties}\label{subsec:elem}

By definition of the OLSMC estimate, the SHCV estimate is exact on the linear span of the control variates \citep{portier2019monte}. Thus, if $f^{(p)}_{\mu,\nu}$ happens to be a polynomial function, then the SHCV estimate will be exact as soon as the maximal degree of the spherical harmonics is equal to or larger than the degree of $f^{(p)}_{\mu,\nu}$. 

\begin{proposition}[Exact rule]\label{prop:exact}
If $\fmn$ is a polynomial function with degree $m$, considering the SHCV estimate and control variates $\varphi =(\varphi_{j})_{j =  1} ^{s_{L,d}}$, if $2L\geq m$ and $n>s_{L,d}$ then $\mathrm{SHCV}$ is exact, i.e., $\mathrm{SHCV}_{n,L}^p(\mu,\nu) = \slice_p^p(\mu,\nu)$.
\end{proposition}

In the Gaussian case of \Cref{remark:GaussSW}, for two proportional  dispersion matrices $\bA$ and $\bB$, say $\bB = \gamma \bA$ for some $\gamma > 0$, the formula of $f_{\mu,\nu}^{(2)}(\theta) = \Wass_2^2(\theta_{\sharp}^\star \mu,\theta_{\sharp}^\star \nu)$ simplifies to a quadratic homogeneous polynomial. Therefore, the SHCV estimate is exact\footnote{See numerical comparison in \Cref{subsec:check} where the SHCV estimate is the only one achieving exact integration (zero error).} as soon as the spherical harmonics of degree two are included as control variates, since the centered integrand is then contained in the linear space spanned by the control variates. 
Actually, this remains true in case $p = 2$ if the measures are related by an affine transformation.

\begin{corollary}[Affine transform] \label{prop:affine_t}
In case $\mu, \nu \in \mathcal{P}_2(\rset^d)$ are related by $X \sim \mu$ and $\alpha X + b \sim \nu$ where $\alpha \in (0, \infty)$ and $b \in \rset^d$ then the $\mathrm{SHCV}$ estimate is exact.
\end{corollary}

The SHCV estimator with spherical harmonics of maximal degree two or higher is guaranteed to have an asymptotic variance that is bounded above by the lower- and upper-CV methods of \citet{nguyen2023control}. Indeed, the latter two methods are control variate methods, each based on a \textit{single} control variate, which happens to be a quadratic polynomial.

\begin{corollary}[Mean invariance]\label{cor:invariance}
For $\mu, \nu \in \mathcal{P}_2(\rset^d)$, the error of the SHCV method is (exactly) invariant under changes of the mean vectors $\mathrm{m}_{\mu}$ and $\mathrm{m}_{\nu}$ of $\mu$ and $\nu$ respectively.
\end{corollary} 
This result implies that we can, without loss of generality, suppose that $\mu$ and $\nu$ are centered. The reason is that the effect of the two mean vectors being potentially different is the translation term $\inner{\theta}{\mathrm{m}_Y - \mathrm{m}_X}$ coming from the optimal transport map that pushes $\theta^\star_\sharp\mu$ to $\theta^\star_\sharp\nu$, and this term produces an additional quadratic term $\inner{\theta}{\mathrm{m}_Y - \mathrm{m}_X}^2$ in the function $f_{\mu,\nu}^{(2)}$. This additional quadratic polynomial in $\theta$ is integrated exactly by SHCV; its integral is $\|\mathrm{m}_X-\mathrm{m}_Y\|^2/d$. See \Cref{remark:GaussSW} for the Gaussian case. 

\subsection{Asymptotic Error Bound}

We provide a convergence rate in probability\footnote{The notation $R_n = \Oh_{\mathbb{P}}(a_n)$, for random variables $R_n$ and constants $a_n > 0$, means that for every $\delta > 0$ there exists $K_\delta > 0$ such that $\mathbb{P}(|R_n| > K_\delta a_n) < \delta$ for all $n$.}
of the integration error $\SHCV_{n,L}^p(\mu, \nu) - \SW_p^p(\mu, \nu)$ for general probability measures $\mu, \nu \in \mathcal{P}_p(\rset^d)$ and $p \in [1, \infty)$, where $\SW_p(\mu,\nu)$ is the $\slice$ distance with respect to the uniform distribution on the sphere and $\SHCV_{n,L}^p(\mu,\nu)$ is our estimate in Definition~\ref{def:shcv} based on $n$ random projections on the sphere and using spherical harmonics of even degree up to $2L$ as control variates.
The guarantee relies on two properties:
first, a general convergence rate of the integration error of the OLSCV estimate in \citet[Theorem~1]{portier2019monte}, available in \cref{subsec:PS2019},
and second, a bound on the uniform approximation error of a Lipschitz function on the sphere by its orthogonal projection on the space of polynomials \citep[Theorem~2]{Newman1964JacksonsTI}. The latter result applies since the integrand $f_{\mu,\nu}^{(p)}$ in \Cref{eq:integrand} is always at least Lipschitz: by \citet[Theorem~2.4]{han2023sliced}, we have
\begin{equation}
\label{eq:intLip}
    \left| f_{\mu,\nu}^{(p)}(\theta) - f_{\mu,\nu}^{(p)}(\gamma) \right|
    \le M_p(\mu,\nu) \, \| \theta - \gamma \|
\end{equation}
for $\theta,\gamma \in \sphere$, where $M_p(\mu,\nu)$ is an explicit function of $p$ and the $p$th order moments of $\mu$ and $\nu$.

The modulus of continuity of $f \in \mathscr{C}(\sphere)$ is defined for $h>0$ by
\begin{equation*}
\omega(h) = \sup \{ |f(\theta)-f(\gamma) | : \rho(\theta,\gamma) \le h  \}
\end{equation*}
with $\rho(\theta,\gamma) = \arccos\langle\theta,\gamma\rangle$ the length of the arc connecting $\theta,\gamma \in \sphere$.
        
	\begin{theorem}[\citet{Newman1964JacksonsTI}, Theorem~2] \label{lem:approx_err}
    There exists an absolute constant $A > 0$ such that for each dimension $d \ge 2$, each function $f \in \mathscr{C}(\sphere)$ and each degree $L$ there exists a polynomial $P_{f,L}$ on $\sphere$ of degree at most $L$ such that 
    \begin{align*}
    \sup_{\theta \in \sphere} |f(\theta) - P_{f,L}(\theta)| \le A \, \omega(d/L).
    \end{align*}
    \end{theorem}

The polynomial $P_{f,L}$ in Theorem~\ref{lem:approx_err} that yields the uniform approximation is in general not the same as the orthogonal projection $\Pi_{f,L} = \sum_{\ell=0}^L \sum_{k=1}^{N_\ell^d} \hat{f}_{\ell,k} \varphi_{\ell,k}$ of $f$ in $L_2(\sphere)$ on the space of spherical harmonics of degree at most $L$, where we use the notation of Proposition~\ref{prop:approx}. Still, the $L_2(\sphere)$ approximation error of $\Pi_{f,L}$ is bounded by the one of $P_{f,L}$, which is in turn bounded by the uniform approximation error of $P_{f,L}$. Applying this argument to the integrand $f_{\mu,\nu}^{(p)}$, we obtain the following convergence rate in probability of the SHCV estimate.

\begin{theorem}[Convergence rate] \label{th:rate}
Let $d \ge 2$, $p \in [1,\infty)$ and $\mu, \nu \in \mathcal{P}_p(\rset^d)$ be fixed. For any degree sequence $L = L_n$ such that $L = \oh(n^{1/(2(d-1))})$ as $n \to \infty$, the integration error of the SHCV estimate with respect to the uniform distribution on the sphere satisfies
\begin{equation} 
\label{eq:rate}
    \left| \SHCV_{n,L}^p(\mu,\nu) - \SW_p^p(\mu,\nu) \right| 
    = \Oh_{\mathbb{P}}(L^{-1} n^{-1/2}). 
\end{equation}
\end{theorem}

The proof of \cref{th:rate} is given in \cref{subsec:proof:rate}.
Writing $L = n^{1/(2(d-1))} / \ell_n$ where $\ell_n > 0$ diverges to infinity but can do so arbitrarily slowly (for instance logarithmically), the convergence rate in \Cref{eq:rate} becomes 
\[
    \left| \SHCV_{n,L}^p(\mu,\nu) - \SW_p^p(\mu,\nu) \right| =
    \Oh_{\mathbb{P}} ( \ell_n n^{-\frac{1}{2} \frac{d}{d-1}} ).
\]
For $d = 3$, this yields the rate $n^{-3/4 + \oh(1)}$ for the SHCV estimate, in comparison to the Monte Carlo rate $n^{-1/2}$.

If the integrand $f_{\mu,\nu}^{(p)}$ has higher-order derivatives, then a faster convergence rate can be expected because of approximation results in \citet{ragozin1971} improving \cref{lem:approx_err}. This concerns for instance the $\slice$ distance between two Gaussian distributions, for which the integrand in Remark~\ref{remark:GaussSW} has derivatives of all orders. In contrast, for discrete measures, the integrand will not even be continuously differentiable, and the Lipschitz property is the best we can hope for. Both cases are illustrated in Appendix~\ref{subsec:case_study}.

\section{Conclusion} \label{sec:conclusion}

We have developed a novel method for reducing the variance of Monte Carlo estimation of the $\slice$ distance using spherical harmonics as control variates. The excellent practical performance of the SHCV estimate against state-of-the-art baselines is confirmed by theoretical properties and a convergence rate in probability for the integration error.

Using control variates with QMC sequences is usually \textit{not} implemented in the same way as with Monte Carlo sequences \citep{hickernell2005control} and would require a particular treatment which is beyond the scope of this paper. In statistical inference with parametric probability measures, note that SHCV is compatible with the computation of gradient $\nabla_{\phi}\slice_p^p(\mu,\nu_{\phi})$ and can be used for \textit{generalized} $\slice$ flows \citep{kolouri2019generalized}.
While the proposed SHCV estimate focuses on the uniform distribution on $\sphere$, it can be extended to more general probability distributions by combining control variates with importance sampling techniques as in \citet{leluc2022}.

\section*{Acknowledgments}

We thank Rémi Flamary from \'Ecole Polytechnique for insightful discussion.
The work of R. Leluc and A. Dieuleveut was supported by Hi!Paris (FLAG project), and ANR-REDEEM project. A. Zhuman gratefully acknowledges a research grant from the \textit{National Bank of Belgium} and of the Research Council of the UCLouvain.

\section*{Impact Statement}
This paper presents work whose goal is to advance the field of Machine Learning. There are many potential societal consequences of our work, none which we feel must be specifically highlighted here.

\bibliography{ref}
\bibliographystyle{icml2024}

\newpage
\appendix
\onecolumn


\begin{center}
\Large
Supplementary Material: \\ Sliced-Wasserstein Estimation with Spherical Harmonic as Control Variates
\end{center}

Appendix \ref{app:algos} gathers the pseudo-code of all the algorithms used in the experiments while Appendix \ref{app:numerical} presents additional details and results for the numerical experiments. \Cref{app:proofs} gathers the proofs of the theoretical results and \Cref{app:supp_theory} is dedicated to additional results on spherical harmonics with (1) a case-study on the regularity of the integrand $\fmn$ and (2) a numerical check of the exact integration rule.

\section{Algorithms for Sliced-Wasserstein Estimates}
\label{app:algos}

This section presents algorithmic details of the different Sliced-Wasserstein estimates in the numerical experiments: 

\textbullet \ \Cref{alg:SWMC_app} corresponds to the standard Monte Carlo estimate where random projections $\theta_1,\ldots,\theta_n \sim \prob = \mathcal{U}(\mathbb{S}^{d-1})$ are used to compute the Wasserstein distance of projected distributions with $\fmn(\theta) = \Wass_p^p(\theta_{\sharp}^\star \mu,\theta_{\sharp}^\star \nu)$. 

\textbullet \  \Cref{alg:SHCV} describes the proposed SHCV estimate which relies on the OLS procedure with spherical harmonics $\varphi_j$ for $j=1,\ldots,s$ as control variates. 

\textbullet \  \Cref{alg:LCVSW} and \Cref{alg:UCVSW} describe the two estimates of \citet{nguyen2023control} which rely on Gaussian approximation with lower and upper control variates. 

\textbullet \ \Cref{alg:CVNN} presents the control neighbors technique of \citet{leluc2023speeding} and consists in building a control functional estimate with the $1$-nearest neighbor estimate of the integrand $f$. 

\textbullet \ \Cref{alg:SWRQMC} implements a (Randomized) Quasi-Monte Carlo estimate. A low-discrepancy sequence on the unit cube $[0,1]^d$ is mapped on $\mathbb{S}^{d-1}$ to obtain a deterministic QMC point set on the unit sphere. Next, rotations from the orthogonal group are applied to randomize the samples.

\begin{algorithm}[h]
\caption{\textcolor{col_mc}{Sliced Wasserstein Monte Carlo}}\label{alg:SWMC_app}
\begin{algorithmic}[1]
\REQUIRE $\mu, \nu \in \mathcal{P}_p(\rset^d)$, number of random projections $n$
\STATE Sample random projections $\theta_1,\ldots,\theta_n$ uniformly on $\sphere$
\STATE Compute $f_n = (\fmn(\theta_i))_{i=1}^{n}$
\STATE Return average $\textcolor{col_mc}{\text{MC}}_{n} = (\1_n/n)^\top f_n \phantom{(\fmn(\theta_i))_{i=1}^{n}}$
\end{algorithmic}
\end{algorithm}

\begin{algorithm}[h!]
\caption{\textcolor{col_shcv}{Spherical Harmonics Control Variate Estimate}} \label{alg:SHCV}
\begin{algorithmic}[1]
\REQUIRE $\mu, \nu \in \mathcal{P}_p(\rset^d)$, number of projections $n$, spherical harmonics $\varphi =(\varphi_{j})_{j=1}^s$
\STATE Sample random projections $\theta_1,\ldots,\theta_n$ uniformly on $\sphere$
\STATE  Compute $f_n = (\fmn(\theta_i))_{i=1}^{n}$ and $\Phi = (\varphi(\theta_i)^\top)_{i=1}^n$ $\phantom{(\fmn(\theta_i))_{i=1}^{n}}$
\STATE $(\Int_n^{\mathrm{ols}},\beta_n) \in \arg\min_{\alpha,\beta} \|f_n - \alpha \1_n - \Phi \beta \|_2^2$ $\phantom{(\fmn(\theta_i))_{i=1}^{n}}$
\STATE Return $\textcolor{col_shcv}{\mathrm{SHCV}}_{n} = \Int_n^{\mathrm{ols}}$
\end{algorithmic}
\end{algorithm}

\begin{algorithm}[h!]
\caption{\textcolor{col_cv_low}{Lower-CV Sliced Wasserstein} \citep{nguyen2023control}}
\begin{algorithmic}[1]
\label{alg:LCVSW}
\REQUIRE $\mu = \sum_{i=1}^{m_X} \alpha_i \delta_{x_i}$ and $\nu=\sum_{j=1}^{m_Y} \beta_j \delta_{y_j}$, number of random projections $n$
\STATE Sample random projections $\theta_1,\ldots,\theta_n$ uniformly on $\sphere$
\STATE Compute $f_n = (\fmn(\theta_i))_{i=1}^{n}$
\STATE Compute $\textcolor{col_mc}{\text{MC}}_{n} = (\1_n/n)^\top f_n$ 
\STATE Compute $\bar{x} = \sum_{i=1}^{m_X} \alpha_i x_i$, and $\bar{y} = \sum_{j=1}^{m_Y} \beta_j y_j$
\STATE $C=(c_1,\ldots,c_n)$ with $c_k = (\theta_k^\top( \bar{x} -\bar{y}))^2 $
\STATE Compute $b =  \|\bar{x}-\bar{y}\|_2^2/d$
\STATE Compute $\gamma_{\text{low}} = \frac{\langle f_n- \textcolor{col_mc}{\text{MC}}_{n}, C-b \rangle }{\|C-b\|_2^2}$ and control variate $\mathcal{C}=(\1_n^\top (C-b))/n$
\STATE Return $\textcolor{col_cv_low}{\text{CV}_{\text{low}}} = \textcolor{col_mc}{\text{MC}}_{n} - \gamma_{\text{low}} \mathcal{C}$
\end{algorithmic}
\end{algorithm}


\begin{algorithm}[h!]
\caption{\textcolor{col_cv_up}{Upper-CV Sliced-Wasserstein } \citep{nguyen2023control}}
\begin{algorithmic}[1]
\label{alg:UCVSW}
\REQUIRE $\mu = \sum_{i=1}^{m_X} \alpha_i \delta_{x_i}$ and $\nu=\sum_{j=1}^{m_Y} \beta_j \delta_{y_j}$, number of random projections $n$
\STATE Sample random projections $\theta_1,\ldots,\theta_n$ uniformly on $\sphere$
\STATE Compute $f_n = (\fmn(\theta_i))_{i=1}^{n}$
\STATE Compute $\textcolor{col_mc}{\text{MC}}_{n} = (\1_n/n)^\top f_n$ 
\STATE Compute $\bar{x} = \sum_{i=1}^{m_X} \alpha_i x_i$, and $\bar{y} = \sum_{j=1}^{m_Y} \beta_j y_j$
\STATE $C=(c_1,\ldots,c_n)$ with $c_k = (\theta_k^\top(\bar{x}-\bar{y}))^2 + \sum_{i=1}^{m_X} \alpha_i(\theta_k^\top(x_i-\bar{x}))^2 +\sum_{j=1}^{m_Y} \beta_j(\theta_k^\top( y_j-\bar{y}))^2  $
\STATE Compute $b = \left(\|\bar{x}-\bar{y}\|_2^2 + \sum_{i=1}^{n_X} \alpha_i\|(x_i-\bar{x})\|_2^2 + \sum_{j=1}^{m_Y} \beta_j\|(y_j-\bar{y})\|_2^2\right)/d$
\STATE Compute $\gamma_{\text{up}} = \frac{\langle f_n- \textcolor{col_mc}{\text{MC}}_{n}, C-b \rangle }{\|C-b\|_2^2}$ and control variate $\mathcal{C}=(\1_n^\top (C-b))/n$
\STATE Return $\textcolor{col_cv_up}{\text{CV}_{up}} = \textcolor{col_mc}{\text{MC}}_{n} - \gamma_{\text{up}} \mathcal{C}$
\end{algorithmic}
\end{algorithm}


\begin{algorithm}[h!]
\caption{\textcolor{col_cvnn}{Control Neighbors Sliced Wasserstein} \cite{leluc2023speeding}}\label{alg:CVNN}
\begin{algorithmic}[1]
\REQUIRE $\mu, \nu \in \mathcal{P}_p(\rset^d)$, number of random projections $n$, probability $\prob \in \mathcal{P}(\mathbb{S}^{d-1})$
\STATE Sample random projections $\theta_1,\ldots,\theta_n \sim \prob$
\STATE Compute $f_n = (\fmn(\theta_i))_{i=1}^{n}$
\STATE Compute $\textcolor{col_mc}{\text{MC}}_{n} = (\1_n/n)^\top f_n$ 
\STATE Compute nearest neighbor evaluations $\hat{f}_n = (\hat{f}_n^{(1)}(\theta_1),\ldots,\hat{f}_n^{(n)}(\theta_n))$
\STATE Return $\textcolor{col_cvnn}{\text{CVNN}}_{n} = \textcolor{col_mc}{\text{MC}}_{n} - [\1_n^\top \{\hat{f}_n-\mathrm{I}(\hat{f}_n)\}]/n$
\end{algorithmic}
\end{algorithm}


\begin{algorithm}[h!]
\caption{\textcolor{col_halton}{Randomized Quasi-Monte Carlo Sliced Wasserstein} \citep{nguyen2023quasimonte}}\label{alg:SWRQMC}
\begin{algorithmic}[1]
\REQUIRE $\mu, \nu \in \mathcal{P}_p(\rset^d)$, number of projections $n$, $\Phi$ the cdf of $\mathcal{N}(0,1)$
\STATE Generate low-discrepancy sequence $U_1,\ldots,U_n$ on $[0,1]^d$
\STATE Compute Gaussian mapping $v_i = \Phi^{-1}(U_i)/\|\Phi^{-1}(U_i)\|_2$, where the standard normal quantile function $\Phi^{-1}$ is applied component-wise.
\STATE Generate a random rotation $R \sim \mathcal{U}(O_d(\rset))$ where $O_d(\rset)= \{R \in \rset^{d \times d} \mid R^\top R = I_d\}$
\STATE Generate random projections $\theta_i=Rv_i$
\STATE Compute $f_n = (\fmn(\theta_i))_{i=1}^{n}$
\STATE Return $\textcolor{col_halton}{\text{RQMC}}_{n} = (\1_n/n)^\top f_n$
\end{algorithmic}
\end{algorithm}

\textbf{Remarks and Practical Details on the Algorithms.} 

\textbullet \ In \Cref{alg:LCVSW}, with the notation $z = \bar{x}-\bar{y}$, the control variate is the centered version of the function $\theta \mapsto |\theta^\top z|^2$, which has known integral $\int_{\sphere} |\theta^\top z|^2 \diff \unif(\theta) = \| z \|^2 / d = b$ (Step~6 in the algorithm) and where $\unif$ denotes the uniform distribution on $\sphere$. The Lower-CV estimate is thus an OLSMC estimate with control variate $\varphi(\theta) = |\theta^\top z|^2 - \| z \|^2 / d$, a polynomial in $\theta$ of degree two. 

\textbullet \ \Cref{alg:UCVSW} is actually an OLSMC estimate with a single control variate given by the centered version of $\theta \mapsto \theta^\top M \theta$ where $M = M_0+M_X+M_Y$ with $M_0 = (\bar{x} - \bar{y}) (\bar{x} - \bar{y})^\top$, $M_X = \sum_{i=1}^{m_X} \alpha_i (x_i-\bar{x}) (x_i-\bar{x})^\top$ and similarly for $M_Y$. The integral with respect to the uniform distribution on the sphere is $\int_{\sphere} \theta^\top M \theta \, \diff\unif(\theta) = \Tr(M)/d = b$ in Step~6 of the algorithm. The control variate is thus the centered quadratic polynomial $\varphi(\theta) = \theta^\top M \theta - \Tr(M)/d$.

\textbullet \ In Step~4 of \Cref{alg:CVNN}, $\hat{f}_n^{(i)}(\theta_i)$ is a leave-one-out nearest-neighbor estimate computed with the Euclidean metric. In Step~5, computing $\Int(\hat{f}_n)$ requires sampling $M$ additional directions $\theta_i$ from $\prob$ and we set $M=n^{1 + 2/d}$ as recommended by the theory in \citet{leluc2023speeding}.

\textbullet \ In \Cref{alg:SWRQMC}, random orthogonal matrices $R \in O_d(\rset)$ are generated using the function \textsf{ortho\_group} of the Python package \textsf{scipy} \citep{virtanen2020scipy}. This function returns random orthogonal matrices drawn from the Haar distribution using a careful QR decomposition\footnote{QR decomposition refers to the factorization of a matrix $X$ into a product $X=QR$ of an orthonormal matrix $Q$ and an upper triangular matrix $R$.} 
as in \citet{mezzadri2007generate}.

\newpage
\section{Numerical Details and Additional Results}\label{app:numerical}

To assess the finite-sample performance of the SHCV estimate, we first present some synthetic data examples involving the computation of $\slice_2^2(\mu,\nu)$ between multivariate Gaussian distributions. We consider different dimensions $d \in \{3;5;6;10;20\}$ and a varying number of random projections $n \in [10^2;10^4]$. We consider the two cases of \textit{exact} distributions in Section \ref{subsec:exact_app} and \textit{sampled} distributions in Section \ref{subsec:sampled}. 
Then in Section \ref{subsec:app_clouds}, we focus on the Sliced-Wasserstein distance $\slice_2(\mu_m,\nu_m)$ between 3D point clouds from the \textsf{ShapeNetCore} dataset of \citet{chang2015shapenet}. In both cases, the true value of the Sliced-Wasserstein distance is estimated with a very large number $n=10^8$ of projections. The different results report the mean squared error $\expec[|{\slice}_{2,n}^2(\mu,\nu)-\slice_2^2(\mu,\nu)|^2]$ for the different estimates $\slice_{2,n}^2(\mu,\nu)$ where the expectation is computed as an average over $100$ independent runs. The experiments were performed on a laptop Intel Core i7-10510U CPU \@ 1.80GHz $\times$ 8. 

\textbf{Methods in competition.} Along with the standard Monte Carlo estimate (\Cref{alg:SWMC_app}) and the proposed SHCV estimate (\Cref{alg:SHCV}), we consider three other baselines based on control variates: the two estimates of \citet{nguyen2023control} based on lower and upper bounds of a Gaussian approximation (\Cref{alg:LCVSW} and \Cref{alg:UCVSW}) and the \textit{control neighbors} estimate of \citet{leluc2023speeding} based on nearest neighbors estimates acting as control variates (\Cref{alg:CVNN}). For completeness, we also include a comparison with (Randomized) Quasi Monte Carlo procedures as in \citet{nguyen2023quasimonte} where deterministic QMC sets from $[0,1]^d$ are mapped to the unit sphere and randomly rotated (\Cref{alg:SWRQMC}). The methods in competition are: the standard Monte Carlo estimate (\textcolor{col_mc}{MC}), the lower-CV (\textcolor{col_cv_low}{$\text{CV}_{low}$}) and upper-CV (\textcolor{col_cv_up}{$\text{CV}_{up}$}) estimates, the control neighbors estimate (\textcolor{col_cvnn}{CVNN}), the proposed estimate with spherical harmonics as control variates (\textcolor{col_shcv}{SHCV}), the Quasi Monte Carlo (\textcolor{col_sobol}{QMC}) and Random Quasi Monte Carlo estimates (\textcolor{col_halton}{RQMC}).

\subsection{Multivariate Gaussian: Exact Case with Distributions $\mu, \nu$ and Smooth Integrands}
\label{subsec:exact_app}

\begin{figure}[h!]
  \centering
  \subfigure[$d=3$]{
  \includegraphics[scale=0.426]{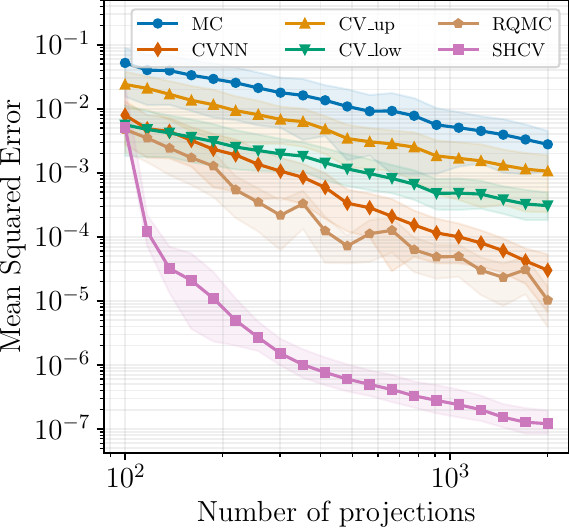}\label{fig:d3_exact_app}}
  \subfigure[$d=6$]{
  \includegraphics[scale=0.426]{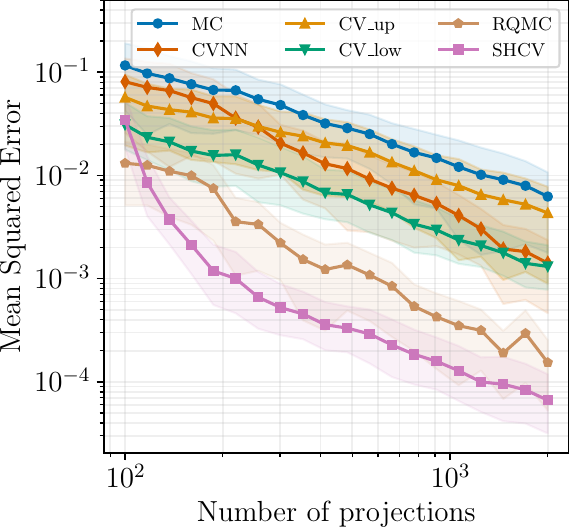}\label{fig:d6_exact_app}}
   \subfigure[$d=3$]{
  \includegraphics[scale=0.426]{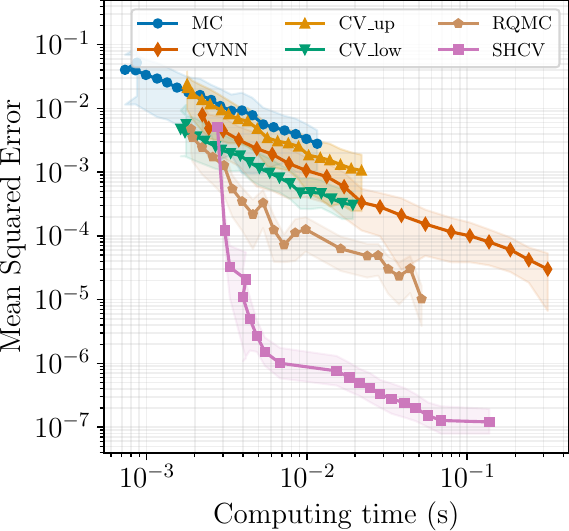}\label{fig:d3_exact_time_app}}
  \subfigure[$d=6$]{
  \includegraphics[scale=0.426]{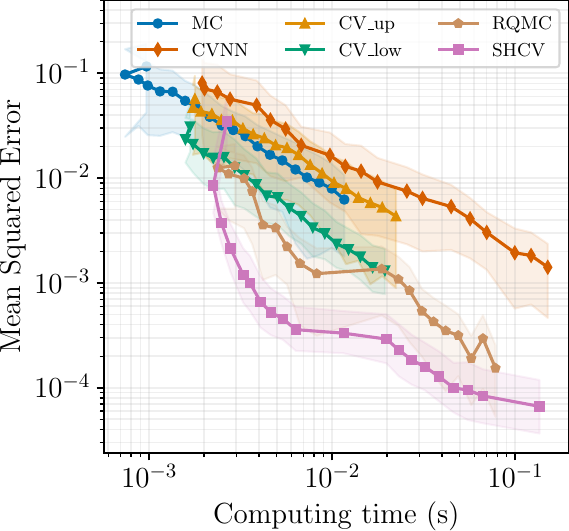}\label{fig:d6_exact_time_app}}
\vspace{-1em}
\caption{MSE and computing time for Gaussian distributions, dimension $d \in \{3;6\}$, obtained over $100$ replications.}
\vspace{-1em}
    \label{fig:synthetic_exact_appendix}
\end{figure}

We compute $\slice_2^2(\mu,\nu)$ between  $\mu=\Normal(a,\mathbf{A})$ and $\nu=\Normal(b,\mathbf{B})$ with means $a,b \sim \Normal(\1_d,I_d)$ and covariance matrices $\mathbf{A}=\Sigma_a^{}\Sigma_a^\top$ and $\mathbf{B}=\Sigma_b^{}\Sigma_b^\top$ where all the entries of $\Sigma_a, \Sigma_b$ are drawn according to $\Normal(0,1)$. The integrand of interest is $f_{\mu,\nu}^{(2)}(\theta)  = |\theta^\top(a-b)|^2 + \bigl(\sqrt{\theta^\top \bA \theta}- \sqrt{\theta^\top \bB \theta}\bigr)^2$. \Cref{fig:synthetic_exact_appendix} reports the MSE of the different methods with respect to both the number of projections and the computing time in dimension $d\in \{3;6\}$, while \Cref{tab:res_gaus_dim_exact_appendix} reports the MSE and computing times (in ms) in dimension $d\in \{5;10;20\}$.

\begin{table}[h]
  \centering
  \begin{tabular}{lcccccc}
     \toprule
    \multirow{2}{2em}{Method} & \multicolumn{2}{c}{$d=5$}  & \multicolumn{2}{c}{$d=10$} & \multicolumn{2}{c}{$d=20$} \\
    \cmidrule(l){2-3}\cmidrule(l){4-5}\cmidrule(l){6-7}
    & MSE & Time & MSE & Time & MSE & Time \\
    \midrule
    \textcolor{col_mc}{MC} & $2.6$e-$2$ & $06.4\pm1.1$ & $8.4$e-$3$ & $06.7\pm0.8$ & $2.0$e-$2$ & $06.6\pm0.5$   \\
    \textcolor{col_cv_low}{$\text{CV}_{low}$} & $4.7$e-$3$ & $11.1\pm3.4$ & $3.0$e-$3$ & $11.6\pm3.0$ & $4.4$e-$3$ & $12.1\pm2.6$ \\
    \textcolor{col_cv_up}{$\text{CV}_{up}$} & $1.5$e-$2$ & $11.4\pm3.5$ & $7.0$e-$3$ & $11.8\pm3.1$ & $1.6$e-$2$ & $13.2\pm2.8$ \\
    \textcolor{col_cvnn}{CVNN} & $7.7$e-$3$ & $43.4\pm13\phantom{.}$ &  $1.0$e-$2$ & $60.0\pm16\phantom{.}$ & $2.6$e-$2$ & $70.6\pm14\phantom{.}$  \\
    \textcolor{col_sobol}{QMC} & $4.9$e-$3$ & $22.4\pm3.4$ & $4.1$e-$3$ & $41.1\pm11\phantom{.}$ & $1.3$e-$2$ & $60.9\pm7.1$  \\
    \textcolor{col_halton}{RQMC} & $7.5$e-$4$ & $25.6\pm6.8$ & $2.4$e-$3$ & $40.9\pm11\phantom{.}$ & $1.4$e-$2$ & $66.4\pm14\phantom{.}$ \\
    \textcolor{col_shcv}{SHCV} & $\mathbf{4.8}$e-$\mathbf{5}$ & $09.8\pm5.1$ & $\mathbf{1.7}$e-$\mathbf{3}$ & $08.9\pm1.2$ & $\mathbf{2.7}$e-$\mathbf{3}$ & $14.3\pm2.0$ \\
    \bottomrule
  \end{tabular}
  \caption{Mean Squared Error (MSE) and computing time (ms) for Gaussian distributions in dimension $d \in \{5;10;20\}$ and $n=500$.}
  \label{tab:res_gaus_dim_exact_appendix}
\end{table}

\subsection{Multivariate Gaussian: Sampled Case with Discrete Distributions $\mu_m, \nu_m$} \label{subsec:sampled}

\begin{figure}[h!]
  \centering
  \subfigure[$d=3$]{
  \includegraphics[scale=0.426]{graphs/synthetic_d3_samples_errors.pdf}\label{fig:d3_sampled_app}}
  \subfigure[$d=6$]{
  \includegraphics[scale=0.426]{graphs/synthetic_d6_samples_errors.pdf}\label{fig:d6_sampled_app}}
  \subfigure[$d=3$]{
  \includegraphics[scale=0.426]{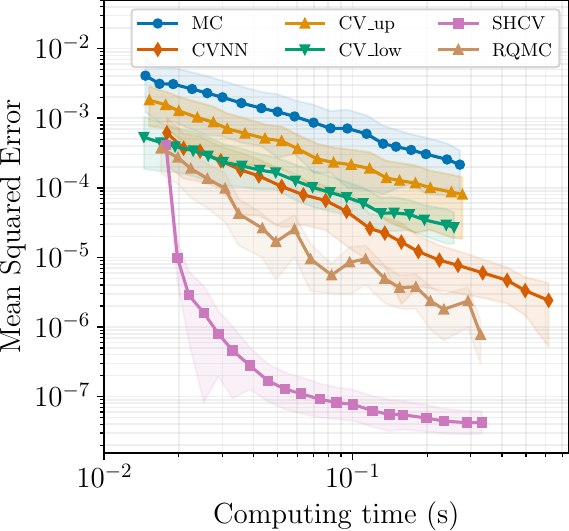}\label{fig:d3_sampled_time}}
  \subfigure[$d=6$]{
  \includegraphics[scale=0.426]{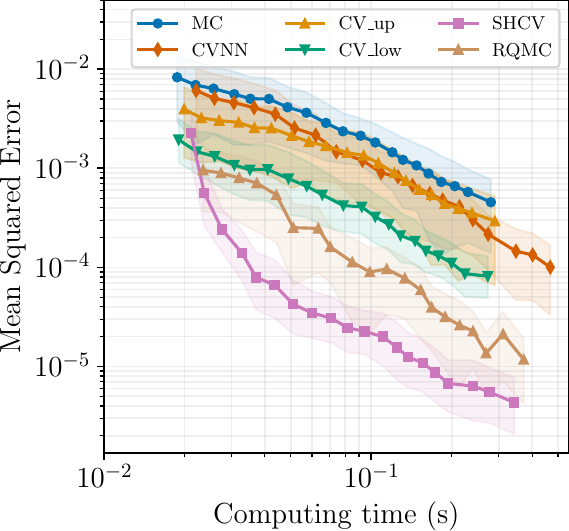}\label{fig:d6_sampled_time}}
  \vspace{-0.2cm}
\caption{MSE for sampled Gaussian distributions supported on $m=1000$ points, dimension $d \in \{3;6\}$, obtained over $100$ replications.}
    \label{fig:synthetic_sampled_appendix}
\end{figure}

We compute $\slice_2^2(\mu_m, \nu_m)$ between empirical distributions  $\mu_m= m^{-1}\sum_{i=1}^m \delta_{x_i}$ and $\nu_m = m^{-1}\sum_{j=1}^m \delta_{y_j}$,
\begin{align*}
     x_1,\ldots,x_m &\sim \mu=\Normal(a,\mathbf{A}), \quad y_1,\ldots,y_m \sim \nu=\Normal(b,\mathbf{B})
\end{align*}
with means $a,b \sim \Normal_d(\1_d,I_d)$ and covariance matrices $\mathbf{A}=\Sigma_a^{}\Sigma_a^\top$ and $\mathbf{B}=\Sigma_b^{}\Sigma_b^\top$ where all the entries of $\Sigma_a, \Sigma_b$ are drawn according to $\Normal(0,1)$. This time, instead of evaluating the integrand $f_{\mu,\nu}^{(2)}$ directly, we assume that one only has access to data samples from the source and the target. We use $m=1000$ data samples from the source $x_1,\ldots,x_m \sim \mu$ and the target $y_1,\ldots,y_m \sim \nu$. \Cref{fig:synthetic_sampled_appendix} reports the mean squared errors of the different methods with respect to both the number of projections and the computing time in dimension $d\in \{3;6\}$.
Tables \ref{tab:synthetic_sampled_mse} and \ref{tab:synthetic_sampled_times} report the MSE and computing times (in ms) in dimension $d\in \{5;10;20\}$.  
\vspace{-1em}
\begin{table}[h!]
  \centering
  \begin{tabular}{lcccccc}
    \toprule
    \multirow{2}{2em}{Method} & \multicolumn{2}{c}{$d=5$}  & \multicolumn{2}{c}{$d=10$} & \multicolumn{2}{c}{$d=20$} \\
    \cmidrule(l){2-3}\cmidrule(l){4-5}\cmidrule(l){6-7}
    & $n=500$ & $n=1000$ & $n=500$ & $n=1000$ & $n=500$ & $n=1000$ \\
    \midrule
     \textcolor{col_mc}{MC} & $1.45$e-$3$ & $8.05$e-$4$ & $9.45$e-$4$ & $5.46$e-$4$ & $1.47$e-$3$ & $6.67$e-$4$  \\
    \textcolor{col_cv_low}{$\text{CV}_{\text{low}}$} & $2.67$e-$4$ & $1.46$e-$4$ & $3.45$e-$4$ & $1.63$e-$4$ & $3.82$e-$4$ & $1.73$e-$4$ \\
    \textcolor{col_cv_up}{$\text{CV}_{\text{up}}$}&  $8.44$e-$4$ & $4.82$e-$4$  & $7.51$e-$4$ & $4.76$e-$4$ & $1.09$e-$3$ & $4.73$e-$4$   \\
    \textcolor{col_cvnn}{CVNN}  & $4.29$e-$4$ & $1.79$e-$4$ & $1.12$e-$3$ & $4.77$e-$4$ & $2.14$e-$3$ & $1.44$e-$3$ \\
    \textcolor{col_sobol}{QMC}  & $2.91$e-$4$ & $7.97$e-$5$ & $2.37$e-$4$ & $8.35$e-$5$ & $6.60$e-$4$ & $1.42$e-$4$ \\
    \textcolor{col_halton}{RQMC}  & $5.80$e-$5$ & $1.91$e-$5$ & $2.75$e-$4$ & $1.20$e-$4$ & $1.17$e-$3$ & $3.64$e-$4$  \\
    \textcolor{col_shcv}{SHCV} & $\mathbf{2.68}$e-$\mathbf{6}$ & $\mathbf{5.63}$e-$\mathbf{7}$ & $\mathbf{1.93}$e-$\mathbf{4}$ & $\mathbf{6.86}$e-$\mathbf{5}$ & $\mathbf{2.95}$e-$\mathbf{4}$ & $\mathbf{1.16}$e-$\mathbf{4}$  \\
    \bottomrule
  \end{tabular}
  \caption{Mean Squared Error (MSE) for sampled Gaussian distributions supported on $m=1000$ points in dimension $d \in \{5;10;20\}$ with $n \in \{500;1000\}$ projections, obtained over $100$ independent runs.}
  \label{tab:synthetic_sampled_mse}
\end{table}

\begin{table}[h!]
  \centering
  \begin{tabular}{lcccccc}
    \hline
     \toprule
    \multirow{2}{2em}{Method} & \multicolumn{2}{c}{$d=5$}  & \multicolumn{2}{c}{$d=10$} & \multicolumn{2}{c}{$d=20$} \\
    \cmidrule(l){2-3}\cmidrule(l){4-5}\cmidrule(l){6-7}
    & $n=500$ & $n=1000$ & $n=500$ & $n=1000$ & $n=500$ & $n=1000$ \\
    \midrule
    \textcolor{col_mc}{MC} & $81.1\pm3.5$ & $152\pm0.9$ & $80.7\pm4.4$& $152\pm1.2$ & $81.1\pm1.8$ &  $155\pm2.3$  \\
    \textcolor{col_cv_low}{$\text{CV}_{\text{low}}$}  & $79.7\pm1.1$ & $151\pm1.1$ & $80.1\pm1.4$ & $150\pm2.2$ & $80.0\pm1.0$ & $155\pm4.0$  \\
    \textcolor{col_cv_up}{$\text{CV}_{\text{up}}$} & $83.0\pm1.2$ & $157\pm1.7$ & $83.0\pm1.7$ & $157\pm1.6$ & $83.1\pm1.5$ & $160\pm3.0$  \\
    \textcolor{col_cvnn}{CVNN}  & $110\phantom{.}\pm2.2$& $208\phantom{}\pm1.5$& $122\phantom{.}\pm1.6$ & $244\phantom{}\pm2.1$ & $127\phantom{.}\pm1.4$& $275\phantom{}\pm3.6$  \\
    \textcolor{col_sobol}{QMC}  & $100\phantom{.}\pm1.2$ & $174\pm1.3$ & $113\phantom{.}\pm1.4$ & $187\phantom{.}\pm1.0$ & $129\phantom{.}\pm1.4$ & $214\phantom{.}\pm3.5$ \\
    \textcolor{col_halton}{RQMC}  & $96.3\pm2.2$ & $190\pm1.4$ & $113\phantom{.}\pm1.2$ & $188\phantom{.}\pm2.4$& $130\phantom{.}\pm1.0$ & $214\phantom{.}\pm3.8$  \\
    \textcolor{col_shcv}{SHCV} & $89.0\pm6.3$ & $175\pm3.4$ & $89.0\pm4.5$ & $176\pm2.4$ & $88.1\pm2.8$ & $180\pm5.6$  \\
    \bottomrule
  \end{tabular}
  \caption{Computing time (ms) for sampled Gaussian distributions supported on $m=1000$ points in dimension $d \in \{5;10;20\}$ with $n \in \{500;1000\}$ projections, obtained over $100$ independent runs.}
  \label{tab:synthetic_sampled_times}
\end{table}

\subsection{Empirical Distributions of $3$D Point Clouds} \label{subsec:app_clouds}

\begin{figure}[h!]
  \centering
  \subfigure[plane]{
  \includegraphics[scale=0.4]{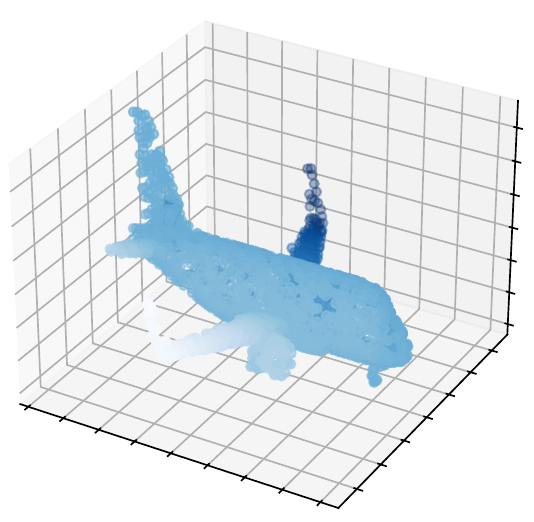}\label{fig:plane}}
  \hspace{1cm}
   \subfigure[lamp]{
  \includegraphics[scale=0.4]
  {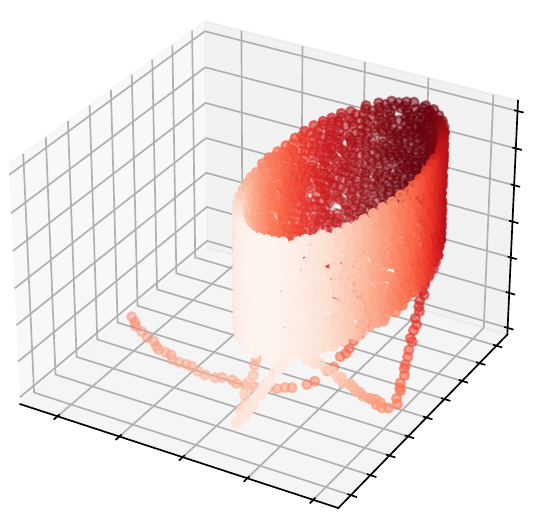}\label{fig:lamp}}
  \hspace{1cm}
  \subfigure[bed]{
  \includegraphics[scale=0.4]
  {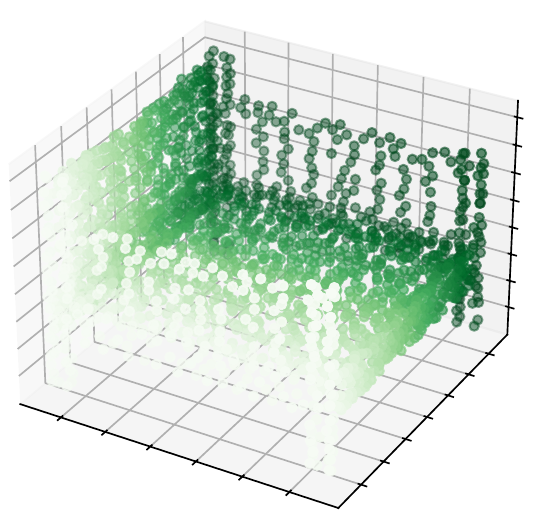}\label{fig:bed}}
  \vspace{-0.2cm}
\caption{Visualization of $3$D point clouds of \textsf{plane, lamp, bed}.}
    \label{fig:visu_clouds}
\end{figure}

We use three point clouds taken at random from the \textsf{ShapeNetCore} dataset of \citet{chang2015shapenet} corresponding to the objects \textsf{plane}, \textsf{lamp} and \textsf{bed}, each composed of $m=2048$ points of $\rset^3$ (see Figure \ref{fig:visu_clouds} below for visualization). The SHCV estimate uses spherical harmonics up to degree $2L=16$ leading to $s=152$ control variates. \Cref{fig:box} reports the boxplots of the error distribution 
$(\widehat{\slice}_n(\mu_m,\nu_m)-\slice(\mu_m,\nu_m))$ for the different $\slice$ estimates based on $n$ random projections with $n \in \{100;250;500;1000\}$ obtained over $100$ independent runs while \Cref{tab:mse_clouds} presents the mean squared errors $\expec[|\widehat{\slice}_n(\mu_m,\nu_m)-\slice(\mu_m,\nu_m)|^2]$ for the different $\slice$ estimates with $n \in \{100;500;1000\}$. Once again, our SHCV estimate provides the best performance with huge gains in terms of variance reduction.

\begin{table}[h!]
  \centering
  \begin{tabular}{lccccccccc}
    \toprule
    \multirow{2}{3em}{\centering Method} & \multicolumn{3}{c}{\textsc{plane/lamp}} & \multicolumn{3}{c}{\textsc{lamp/bed}} & \multicolumn{3}{c}{\textsc{plane/bed}} \\
    \cmidrule(l){2-4}\cmidrule(l){5-7}\cmidrule(l){8-10}
    & $n=100$ & $n=500$ &  $n=1000$ &  $n=100$ & $n=500$ &  $n=1000$ & $n=100$ & $n=500$ &  $n=1000$ \\
    \midrule
     \textcolor{col_mc}{MC} & $5.1$e-$5$ & $1.1$e-$5$ & $6.0$e-$6$ & $2.0$e-$4$ & $3.8$e-$5$ & $1.5$e-$5$  & $1.3$e-$4$ & $2.6$e-$5$ & $1.2$e-$5$\\
    \textcolor{col_cv_low}{$\text{CV}_{\text{low}}$} & $4.8$e-$5$ & $9.3$e-$6$ & $5.1$e-$6$ & $9.3$e-$5$ & $1.4$e-$5$ & $6.5$e-$6$ & $4.4$e-$5$ & $9.8$e-$6$ & $5.2$e-$6$\\
    \textcolor{col_cv_up}{$\text{CV}_{\text{up}}$}&  $1.1$e-$5$ & $2.2$e-$6$  & $1.3$e-$6$ & $1.1$e-$5$ & $2.3$e-$6$ & $1.0$e-$6$  & $6.6$e-$5$ & $1.5$e-$5$ & $8.7$e-$6$ \\
    \textcolor{col_cvnn}{CVNN}  & $1.3$e-$5$ & $5.2$e-$7$ & $1.8$e-$7$ & $2.4$e-$5$ & $1.1$e-$6$ & $3.5$e-$7$ & $3.2$e-$5$ & $1.1$e-$6$ & $3.3$e-$7$ \\
    \textcolor{col_sobol}{QMC}  & $1.4$e-$5$ & $6.9$e-$7$ & $2.8$e-$8$ & $2.4$e-$5$ & $3.8$e-$7$ & $2.9$e-$7$ & $1.2$e-$4$ & $1.2$e-$7$ & $4.9$e-$7$ \\
    \textcolor{col_halton}{RQMC}  & $7.7$e-$6$ & $2.2$e-$7$ & $8.1$e-$8$ & $1.5$e-$5$ & $2.7$e-$7$ & $1.3$e-$7$  & $2.6$e-$5$ & $1.0$e-$6$ & $3.0$e-$7$ \\
    \textcolor{col_shcv}{SHCV} & $\mathbf{5.3}$e-$\mathbf{6}$ & $\mathbf{2.1}$e-$\mathbf{8}$ & $\mathbf{2.1}$e-$\mathbf{8}$ & $\mathbf{2.5}$e-$\mathbf{6}$ & $\mathbf{1.5}$e-$\mathbf{8}$ & $\mathbf{1.4}$e-$\mathbf{8}$ & 
    $\mathbf{1.3}$e-$\mathbf{5}$ &
    $\mathbf{3.7}$e-$\mathbf{8}$ &
    $\mathbf{3.6}$e-$\mathbf{8}$ \\
    \bottomrule
  \end{tabular}
  \caption{Mean Squared Error $\expec[|\widehat{\slice}_n(\mu_m,\nu_m)-\slice(\mu_m,\nu_m)|^2]$ for different SW estimates based on $n$ random projections with $n \in \{100;500;1000\}$ obtained over $100$ independent runs.}
  \label{tab:mse_clouds}
\end{table}

\begin{table}[h!]
  \centering
  \begin{tabular}{lccccccccc}
    \toprule
    \multirow{2}{3em}{\centering Method} & \multicolumn{3}{c}{\textsc{plane/lamp}} & \multicolumn{3}{c}{\textsc{lamp/bed}} & \multicolumn{3}{c}{\textsc{plane/bed}} \\
    \cmidrule(l){2-4}\cmidrule(l){5-7}\cmidrule(l){8-10}
    & $n=100$ & $n=500$ &  $n=1000$ &  $n=100$ & $n=500$ &  $n=1000$ & $n=100$ & $n=500$ &  $n=1000$ \\
    \midrule
     \textcolor{col_mc}{MC} & 
     $36.4\pm10$ & $184\pm3$ & $309\pm3$ & 
     $37.9\pm10$ & $184\pm3$ & $308\pm3$ &
     $35.6\pm10$ & $185\pm3$ & $308\pm3$\\
    \textcolor{col_cv_low}{$\text{CV}_{\text{low}}$} & $36.8\pm10$ & $180\pm1$ & $301\pm1$ & 
    $38.4\pm10$ & $181\pm2$ & $300\pm2$ &
    $36.1\pm11$ & $181\pm1$ & $300\pm1$\\
    \textcolor{col_cv_up}{$\text{CV}_{\text{up}}$}&  
    $38.1\pm10$ & $187\pm2$ & $316\pm1$ & 
    $39.7\pm10$ & $187\pm1$ & $315\pm1$ &
    $37.2\pm11$ & $188\pm1$ & $314\pm1$ \\
    \textcolor{col_cvnn}{CVNN}  & 
    $40.4\pm11$ & $220\pm1$ & $417\pm1$ & 
    $42.0\pm11$ & $220\pm1$ & $414\pm2$ &
    $39.5\pm11$ & $221\pm1$ & $414\pm1$ \\
    \textcolor{col_sobol}{QMC}  & 
    $32.9\pm12$ & $194\pm3$ & $324\pm1$ & 
    $27.9\pm10$ & $194\pm2$ & $327\pm1$ &
    $38.0\pm13$ & $195\pm2$ & $325\pm1$ \\
    \textcolor{col_halton}{RQMC}  &  
    $40.6\pm11$ & $202\pm2$ & $346\pm2$ & 
    $42.0\pm10$ & $203\pm4$ & $344\pm2$  & 
    $39.3\pm11$ & $202\pm1$ & $343\pm1$ \\
    \textcolor{col_shcv}{SHCV} &  
    $37.9\pm10$ & $185\pm2$ & $311\pm1$ &  
    $39.7\pm10$ & $185\pm1$ & $310\pm1$ &  
    $37.1\pm11$ & $186\pm1$ & $310\pm1$\\
    \bottomrule
  \end{tabular}
  \caption{Computing times (ms) for different SW estimates based on $n$ random projections with $n \in \{100;500;1000\}$ obtained over $100$ independent runs.}
  \label{tab:times_clouds}
\end{table}

\subsection{Kernel Support Vector Machines and Sliced-Wasserstein Distances}
\label{subsec:kernel_svm}

\textbf{Kernel Support Vector Machines.} Kernel Support Vector Machines (SVMs) are a powerful class of supervised learning algorithms used for classification and regression tasks. At their core, SVMs aim to find a hyperplane that best separates the data into different classes while maximizing the margin between these classes. However, in cases where the classes are not linearly separable, Kernel SVMs come into play. The key idea behind Kernel SVMs is to map the input data $x \in \rset^q$ into a higher-dimensional space via $\phi(x)$, where a linear separation might be achievable (see \Cref{fig:visu_kernels} below).

 \begin{figure}[h!]
  \centering
  \includegraphics[scale=0.22]{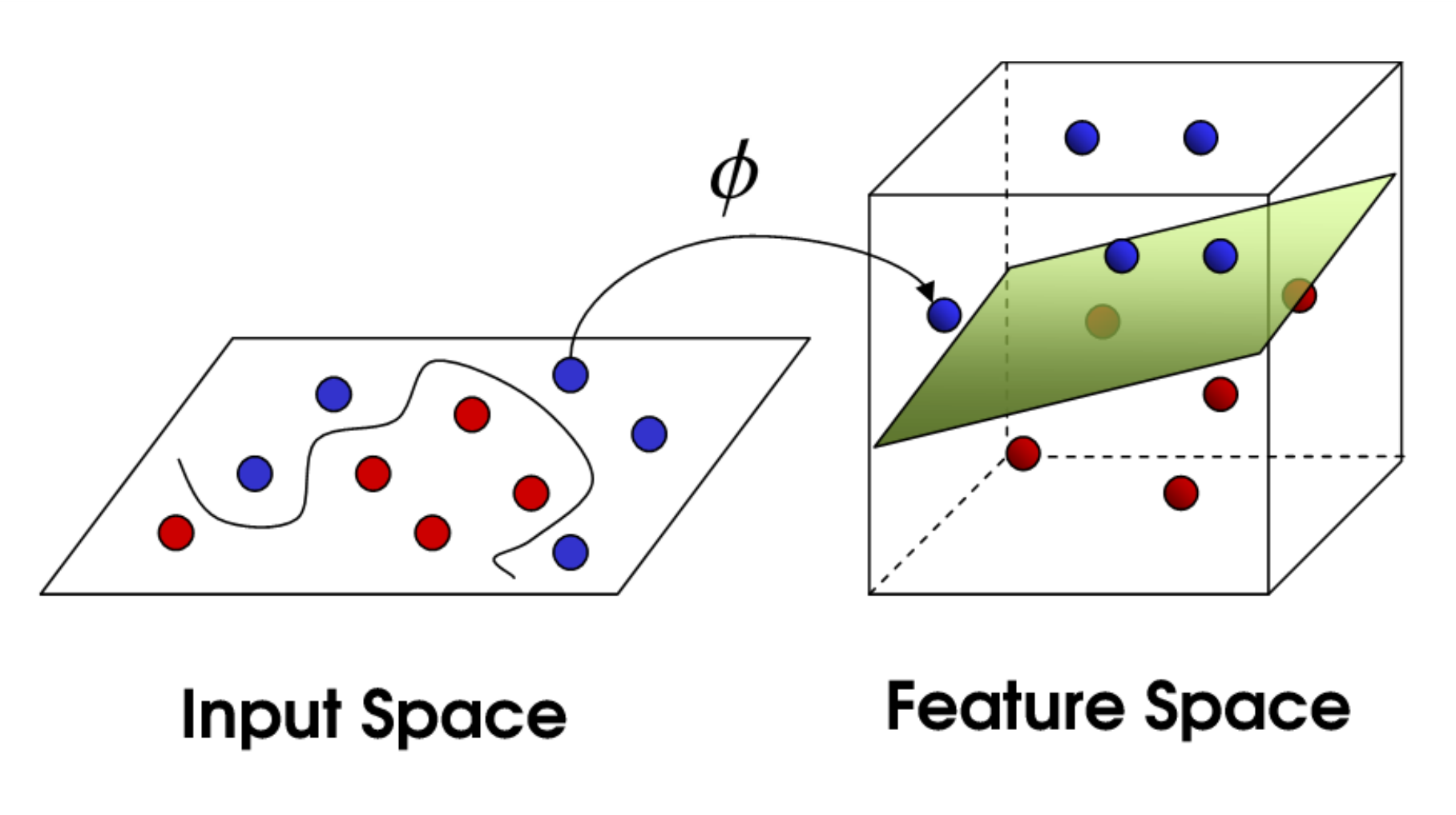}\label{fig:kernel_phi}
\caption{Visualization of Kernel transformation.}
    \label{fig:visu_kernels}
\end{figure}

The kernel function $k(\cdot,\cdot)$ computes the similarity between pairs of data points in the higher-dimensional space without explicitly calculating the transformed data. Given training data $x_i \in \rset^q$ with $i=1,\ldots,n$ divided into two classes and a vector of labels $y \in \{-1;+1\}$, the goal is to find the parameters $\omega \in \rset^q$ and $b \in \rset$ such that the prediction given by $\mathrm{sign}(\omega^\top \phi(x_i) +b)$ is correct for most samples.

In Scikit-learn \citep{pedregosa2011scikit}, the `SVC class' provides an implementation of Kernel SVMs. The objective function (also known as the loss function) for the SVM is derived from the problem of finding the optimal hyperplane that maximizes the margin between different classes. The formulation includes a regularization term $C$ to balance the desire for a large margin with the goal of minimizing classification errors and the optimization problem of interest is given by
\begin{align*}
\min _{\omega, b, \zeta} \frac{1}{2} \omega^\top \omega+C \sum_{i=1}^n \zeta_i \quad
	\text{ subject to } y_i\left(\omega^\top \phi\left(x_i\right)+b\right) \geq 1-\zeta_i 	\text{ and } \zeta_i \geq 0, i=1, \ldots, n.
\end{align*}
The parameter $C$, which is shared among all Support Vector Machine (SVM) kernels, balances the compromise between misclassifying training examples and maintaining a simple decision surface. A lower $C$ value results in a smoother decision surface, while a higher $C$ value strives to accurately classify all training examples. This problem is usually solved with the dual formulation through the Kernel matrix $K(x_i,x_j) = \langle \phi(x_i),\phi(x_j) \rangle$.

 \begin{figure}[H]
  \centering
  \includegraphics[scale=0.6]{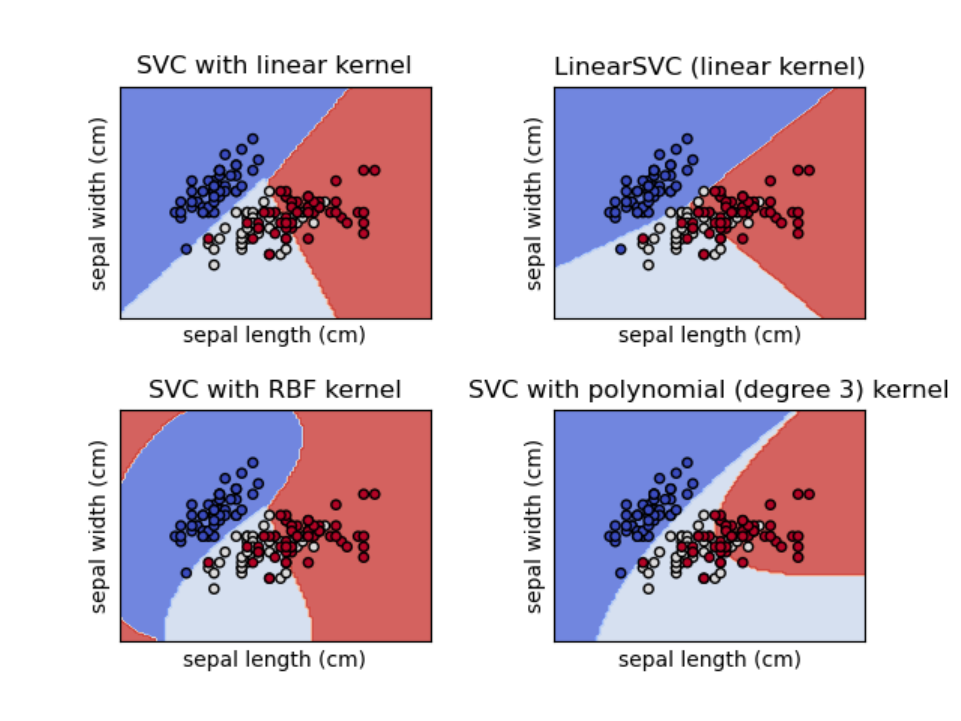}\label{fig:kernel_illustration}
  \vspace{-0.5cm}
\caption{Comparison of different linear SVM classifiers on a 2D projection of the iris dataset.}
    \label{fig:sklearn_kernels}
\end{figure}

\textbf{Sliced-Wasserstein Gaussian Kernels.} SW distances inherit topological properties from the original Wasserstein distance such as the weak convergence of probability measures \citep{nadjahi2020statistical}. 
$\slice$ distances are useful in the context of distribution regression \citep{szabo2016learning} where the goal is to learn a real-valued function defined on the space of probability distributions from a sequence of observations. The classical Wasserstein distance cannot be used to build a positive definite kernel on the space of distributions as the induced Wasserstein space has curvature \citep{feragen2015geodesic}. To overcome this problem, a $\slice$-based kernel for absolutely continuous distributions was first introduced in \citet{kolouri2016sliced} and later on extended to empirical distributions by \citet{meunier2022distribution}. For any $\gamma>0$, both $e^{-\gamma \SW_2^2(\cdot,\cdot)}$ and $e^{-\gamma \SW_1(\cdot,\cdot)}$ are valid kernels on probability measures. 

Using the $\slice$ kernel of \citet{kolouri2016sliced} defined by $k(\mu_i,\mu_j) = \exp(-\gamma \slice_2^2(\mu_i,\mu_j))$ with $\gamma >0$, we consider the image classification task on the \textsf{digits} dataset of \citet{digits_dataset}. This dataset contains $N=1\,797$ images of size $8 \times 8$ and the goal is to predict the label $z \in \{0,1,\ldots,9\}$ of the handwritten digit. Similarly to \citet{meunier2022distribution}, we convert each image to an histogram so that the $\slice$ kernels have probability distributions as inputs. 
We use a train/test split of size $80/20$ giving $N_{\text{train}}=1\,437$ training and $N_{\text{test}}=360$ testing images respectively where each set is balanced. The Kernel SVM classification requires to compute the Gram matrices $K=(k(\mu_i,\mu_j))$ with $(i,j) \in [N_{\text{train}}] \times [N_{\text{train}}]$ and with $(i,j) \in [N_{\text{train}}] \times [N_{\text{test}}]$. Computational benefits occur with the SHCV estimate when many $\slice$ distances need to be computed. Using the projections $\theta_1,\ldots,\theta_n \sim \prob$ and the spherical harmonics $\varphi=(\varphi_j)_{j=1}^s$ to compute $\Pi = \Phi(\Phi^\top \Phi)^{-1} \Phi^\top$ with $\Phi = (\varphi(\theta_i)^\top)_{i=1}^n$, we need to compute the weights $w = [(\Int-\Pi)\1_n]/[\1_n^\top (\Int-\Pi)\1_n]$ \emph{only once} in order to obtain a linear integration rule $w_n^\top f_{n}^{(i,j)}$ with $f_{n}^{(i,j)} = (f_{\mu_i,\nu_j}^{(p)}(\theta_1),\ldots,f_{\mu_i,\nu_j}^{(p)}(\theta_n))^\top$ to estimate $\slice_p^p(\mu_i,\nu_j)$.

\begin{minipage}{0.48\textwidth}
\begin{algorithm}[H]
\caption{\textcolor{col_mc}{SW-Kernel Monte Carlo}}\label{alg:MC_kernel}
\begin{algorithmic}[1]
\REQUIRE $\mathcal{D}_X = (\mu_1,\ldots,\mu_{N_X}), \mathcal{D}_Y = (\nu_1,\ldots,\nu_{N_Y})$, nb random projections $n \in \nset$, $\prob \in \mathcal{P}(\mathbb{S}^{d-1})$
\STATE Initialize Kernel matrix $K \in \rset^{N_X \times N_Y}$
\STATE Sample random projections $\theta_1,\ldots,\theta_n \sim \prob$
\STATE \textbf{For} $i=1,\ldots,N_X$ \textbf{do}
\STATE \quad \textbf{For} $j=1,\ldots,N_Y$ \textbf{do}
\STATE \quad \quad $f_{n}^{(i,j)} = (f_{\mu_i,\nu_j}^{(2)}(\theta_k))_{k=1}^n$
\STATE \quad \quad $\textcolor{col_mc}{\text{MC}}_{n} = (\1_n/n)^\top f_{n}^{(i,j)}$
\STATE \quad \quad $K[i,j] = \exp(-\gamma \textcolor{col_mc}{\text{MC}}_{n})$
\STATE Return Kernel matrix $K$
\end{algorithmic}
\end{algorithm}
\end{minipage}
\begin{minipage}{0.48\textwidth}
\begin{algorithm}[H]
\caption{\textcolor{col_shcv}{SW-Kernel SHCV}}\label{alg:SHCV_Kernel}
\begin{algorithmic}[1]
\REQUIRE $\mathcal{D}_X = (\mu_1,\ldots,\mu_{N_X}), \mathcal{D}_Y = (\nu_1,\ldots,\nu_{N_Y})$, $n \in \nset$, $\prob \in \mathcal{P}(\mathbb{S}^{d-1})$, control variates $\varphi=(\varphi_j)_{j=1}^s$
\STATE Sample random projections $\theta_1,\ldots,\theta_n \sim \prob$
\STATE Compute $\Phi = (\varphi(\theta_i)^\top)_{i=1}^n$ and $\textcolor{col_shcv}{\text{SHCV}}$ weights $w_n$
\STATE \textbf{For} $i=1,\ldots,N_X$ \textbf{do}
\STATE \quad \textbf{For} $j=1,\ldots,N_Y$ \textbf{do}
\STATE \quad \quad $f_{n}^{(i,j)} = (f_{\mu_i,\nu_j}^{(2)}(\theta_k))_{k=1}^n$
\STATE \quad \quad $\textcolor{col_shcv}{\text{SHCV}}_{n} = w_n^\top f_{n}^{(i,j)}$
\STATE \quad \quad $K[i,j] = \exp(-\gamma \textcolor{col_shcv}{\text{SHCV}}_{n})$
\STATE Return Kernel matrix $K$
\end{algorithmic}
\end{algorithm}
\end{minipage}

As mentioned in \Cref{rem:multiple}, computational benefits occur when $K > 1$ integrals need to be computed since the SHCV time complexity scales as $\mathcal{O}(K n \omega_f + \omega(\Phi))$ compared to $\mathcal{O}(K n \omega_f)$ for the standard Monte Carlo estimate when dealing with $K$ integrals. Thus, when $K$ is large, the extra term $\omega(\Phi)$ becomes negligible. In the experiments we are dealing with images and probability measures $\mu_1,\ldots,\mu_{N_X}, \nu_1, \ldots, \nu_{N_Y}$ in dimension $d=2$ with a number of Monte Carlo projections $n \in \{25;50;75;100\}$. The total number of integrals to be computed are:

\textbullet \ $K_{XX}= \binom{N_X}{2}= 1437 \times 1436 / 2 = 1,024,581$ for the 'train/train' kernel matrix. \\
\textbullet \ $K_{XY}=N_X \times N_Y = 1437 \times 360 = 517,320$ for the 'train/test' kernel matrix.

Below, we compare the time (in $\mu s$) to compute the evaluations $f_n^{(i,j)} = (f_{\mu_i,\nu_j}^{(2)}(\theta_k))_{k=1}^n$ for a single pair $(i,j)$ vs. the time to compute the SHCV weights $w_n$. These weights need to be computed only once for all pairs of images. Interestingly, this extra computing time is of the same order are the time required for the evaluations of the integrands so that in the end it becomes negligible as the SHCV time complexity scales as $\mathcal{O}(K n \omega_f + \omega(\Phi))$ compared to $\mathcal{O}(K n \omega_f)$ for the standard Monte Carlo estimate.

\begin{table}[H]
    \centering
    \begin{tabular}{ccc}
    \toprule
    MC projections $n$ &	evaluations $f_n^{(i,j)}$ in $\mathcal{O}(n \omega_f)$ &	SHCV weights $w_n$ in $\mathcal{O}(\omega(\Phi))$ \\
    \midrule
25	& $422\pm36$ &$207\pm243$ \\
50	& $700\pm16$ & $479\pm516$ \\
75	& $945\pm17$ & $761\pm559$ \\
100	& $1182\pm20$ & $1057\pm654$ \\
\bottomrule
    \end{tabular}
    \caption{Comparison of time computations (in $\mu s$) for the evaluations of the integrands vs. SHCV weights.}
    \label{tab:time_kernels}
\end{table}

\section{Mathematical Proofs}
\label{app:proofs}

\subsection{Proofs of  Lemma~\ref{lem:counting} and \Cref{prop:affine_t}}
\textbf{Proof of Lemma~\ref{lem:counting}.} Let $\mathscr{P}^d_\ell$ be the space of homogeneous polynomials of degree $\ell$. Then $\dim \mathscr{P}^d_\ell = \binom{\ell+d-1}{d-1}$, the number of multi-indices of the form $(\alpha_1,\ldots,\alpha_d)$ for nonnegative integer $\alpha_j$ such that $\sum_{k=1}^d \alpha_k = \ell$. From this space, we need to remove the polynomials that are a multiple of $x_1^2 + \cdots + x_d^2$, since, on the sphere, this factor trivial. Any such multiple is of the form $(x_1^2 + \cdots + x_d^2) p(x)$ with $p \in \mathscr{P}^d_{\ell-2}$. For $\ell \ge 2$, we have \citep[Proposition 2.16]{atkinsonhan} $N^d_\ell = \dim \mathscr{P}^d_\ell - \dim \mathscr{P}^d_{\ell-2}$. But then we have $\sum\limits_{\ell = 2, 4, \ldots, 2k} N^d_\ell =  \sum\limits_{\ell = 2, 4, \ldots, 2k} 
	\left[
		\binom{\ell+d-1}{d-1} - \binom{\ell-2+d-1}{d-1}
	\right] 
	= \binom{2k+d-1}{d-1} - 1.$
 
\textbf{Proof of \Cref{prop:affine_t}.} For $\theta \in \sphere$, we have $\inner{\theta}{X} \sim \theta^\star_{\sharp} \mu$ and $\alpha \inner{\theta}{X} + \inner{\theta}{b} \sim \theta^\star_{\sharp} \nu$. The optimal transport map between $\theta^\star_{\sharp} \mu$ and $\theta^\star_{\sharp} \nu$ is $x \mapsto \alpha x + \inner{\theta}{b}$ and
$f_{\mu,\nu}^{(2)}(\theta) 	= \expec_{\mu} [ \inner{\theta}{(1-\alpha)X-b}^2 ]$, which is a quadratic polynomial in $\theta$.
\subsection{Asymptotic Bound for OLSMC}
\label{subsec:PS2019}

Consider the general set-up of \cref{sec:mc_spherical}, in particular the OLSMC estimate in \cref{eq:olsmc}.

\begin{theorem}[\citet{portier2019monte}] \label{th:olsmc_rate}
Let $(\Theta, \mathcal{F}, P)$ be a probability space. 
Let $\varphi=(\varphi_1,\ldots,\varphi_s)^\top \in (L_2(P))^s$ be a vector of orthonormal control variates, where $s = s_n$ depends on the Monte Carlo sample size $n$. Define $q_n(\theta) = \varphi(\theta)^\top\varphi(\theta)$ for $\theta \in \Theta$. If $\sup_{\theta \in \Theta} q(\theta) = \oh(n/s)$ as $n \to \infty$, then $	\Int_n^{\mathrm{ols}}(f) - \Int (f) = \Oh_{\mathbb{P}} (\sigma_n/\sqrt{n}) \text{ as } n \to \infty$, where $\sigma_n^2 = \Int(\varepsilon_n^2)$, $\varepsilon_n = f - \hat{f}_{s}$, and $\hat{f}_s = \esp[f] + \sum_{i=1}^{s} \inner{f}{\varphi_i} \varphi_i$.
\end{theorem}

 
\subsection{Proof of Theorem~\ref{th:rate}}%
\label{subsec:proof:rate}

To apply \Cref{th:olsmc_rate}, we consider $q_n(\theta) = \varphi(\theta)^\top\varphi(\theta)$ for $\theta \in \sphere$. Since $\varphi_1,\ldots,\varphi_s$ are the orthonormal control variates, \vspace{-0.8em}
\begin{align*}
			q_n(\theta) &= \sum_{i=1}^{s} \varphi_{i}^2(\theta) = \sum_{\ell = 2, 4, \ldots, 2L} \sum_{k=1}^{N_{\ell}^d} \varphi^2_{\ell,k} (\theta).  
		 \end{align*}
		 As a corollary to Theorem~2.9 in \citet{atkinsonhan}, we have $\sum_{k=1}^{N_{\ell}^d} \varphi^2_{\ell,k} (\theta) = N_{\ell}^d / |\sphere|$. In view of \cref{lem:counting},
		\begin{equation}
        \label{eq:B}
		q_n(\theta) 
        = |\sphere|^{-1} \sum_{\ell = 2, 4, \ldots, 2L}  N_{\ell}^d 
        = |\sphere|^{-1} s_{L,d}
		\end{equation}
        with $s = s_{L,d} = \binom{2L+d-1}{d-1}-1$.
 Applying Stirling's formula gives $s_{L,d} = \Oh(L^{d-1})$ as $L \to \infty$.
Hence the condition $\sup_{\theta \in \sphere} q(\theta) = \oh(n/s)$ in \cref{th:olsmc_rate} becomes $s_{L,d}^2 = o(n)$, which is fulfilled as soon as
    \begin{equation} \label{eq:cond_rate}
    L = \oh \rbr{n^{1/(2(d-1))}}, \qquad n \to \infty.
    \end{equation}
		Since the function $\theta \mapsto f(\theta) = f_{\mu,\nu}^{(p)}(\theta) = \Wass_p^p(\theta_{\sharp}^\star \mu,\theta_{\sharp}^\star \nu)$ on $\sphere$ is Lipschitz \citep[Theorem~2.4]{han2023sliced}, the modulus of continuity of $f$ evaluated at $h = d/(2L)$ satisfies
		\begin{align} \label{eq:omega1}
		\omega(d/(2L)) 
        \le \Lip_\rho(f) \, d /(2L) 
        \le \Lip_{\|\cdot\|}(f) \, d /(2L),
		\end{align}
		where $\Lip_\rho(f) = \sup_{\theta,\gamma \in \sphere} |f(\theta)-f(\gamma)| / \rho(\theta,\gamma)$ and $\Lip_{\|\cdot\|}(f) = \sup_{\theta,\gamma \in \sphere} |f(\theta)-f(\gamma)| / \|\theta - \gamma\|$. More precisely, from Theorem~2.4 in \citet{han2023sliced}, we know that 
        \begin{equation} 
        \label{eq:Lipbound}
            \Lip_{\|\cdot\|}(f) \le M_p(\mu,\nu) 
            = p 2^{p-1} \max\{M_p(\mu),M_p(\nu)\}^{(p-1)/p} 
            \rbr{(M_p(\mu))^{1/p} + (M_p(\nu))^{1/p}} 
        \end{equation}
        with $M_p(\mu) = \int_{\rset^d} \|x\|^p \, \diff \mu(x)$ and $M_p(\nu) = \int_{\rset^d} \|x\|^p \, \diff \nu(x)$.
            Hence, by \Cref{lem:approx_err}, the residual function $\varepsilon_n = f - \Pi_{f,L}$ with $\Pi_{f,L} = \sum_{\ell = 0}^L \sum_{k=1}^{N_{2\ell}^d} \inner{f}{\varphi_{2\ell,k}} \varphi_{2\ell,k}$ satisfies
            \begin{equation} \label{eq:varepsilon_inf}
            \sigma_n 
            = \|\varepsilon_n\|_2 
            \le \left\| f -  P_{f,L} \right\|_2 
            \le \left\| f -  P_{f,L} \right\|_\infty
            \le \frac{A M_p(\mu,\nu) d }{2L}.
            \end{equation}
	 	Therefore, by \Cref{th:olsmc_rate} along with~\eqref{eq:cond_rate} and~\eqref{eq:varepsilon_inf},  
		we have, since $\Int(f_{\mu,\nu}^{(p)}) = \SW_p^p(\mu,\nu)$, the rate
		\begin{equation*}
            \left| \SHCV_{n,L}^p(\mu,\nu) - \SW_p^p(\mu,\nu) \right| 
    = \Oh_{\mathbb{P}}(L^{-1} n^{-1/2}). 
            \end{equation*}
\section{Additional Results on Spherical Harmonics}
\label{app:supp_theory}

\subsection{Spherical Harmonics in Small Dimensions}\label{subsec:spherical_small_dim}

\textbf{Circular Harmonics on $\mathbb{S}^1$.} On the unit circle, the spherical harmonics are often called \textit{circular harmonics} and correspond to the Fourier basis defined in circular coordinates $(x,y)=(\cos(t),\sin(t))$ by $\psi_{0}(\theta)=1/\sqrt{2\pi}$ and for $k>0$
\begin{align*}
\psi_{2k}(\theta)= \frac{\sin(kt)}{\sqrt{\pi}}, \qquad 
\psi_{2k-1}(\theta) = \frac{\cos((k-1)t)}{\sqrt{\pi}}.
\end{align*}
The circular harmonics are $\varphi_{1,0} = \psi_0$ and for $\ell \geq 1$, $\varphi_{\ell,1} = \psi_{2\ell-1}$ and $\varphi_{\ell,2} = \psi_{2\ell}$.

\textbf{Spherical Harmonics on $\mathbb{S}^2$.} On the sphere $\mathbb{S}^2$ there exist $N_{\ell}^3 = 2\ell + 1$ linearly independent complex spherical harmonics $\psi_{\ell,k}: \mathbb{S}^2 \to \mathbb{C}$ of degree $\ell$. Using the spherical coordinates $(x,y,z)=(\cos(u)\sin(v), \sin(u)\sin(v), \cos(v))$ with $0 \leq u < 2\pi$ and $0 \leq v < \pi$, the Laplace spherical harmonics are generally defined for all $-\ell\leq k \leq \ell$ by
\begin{align*}
\psi_{\ell,k}(u,v) = \sqrt{\frac{(2\ell + 1)(\ell-k)!}{4\pi  (\ell + k)!}} P_{\ell}^k(\cos(v)) e^{iku},
\end{align*}
where $P_{\ell}^k$ is the associated Legendre polynomial defined by $P_{\ell}^k(x) = (-1)^k (1-x^2)^{k/2} (\diff^k/\diff x^k) P_{\ell}(x)$ with $P_\ell$ the Legendre polynomial of degree $\ell$ and $P_{\ell}^{-k}(x) = (-1)^k [(\ell-k)!/(\ell+k)!] P_{\ell}^k(x)$. This definition ensures orthonormality $\int_{v=0}^\pi \int_{u=0}^{2\pi} \psi_{\ell,k} \psi_{\ell',k'}^{\star} \diff \Omega = \delta_{\ell,\ell'}\delta_{k,k'}$ with $\diff \Omega = \sin(v) \diff u \diff v$. A real basis of spherical harmonics $\varphi_{\ell,k}: \mathbb{S}^2 \to \rset$ can be defined in terms of their complex analogues by setting $\varphi_{\ell,0}=\psi_{\ell,0}$ and taking the real and imaginary parts as $\varphi_{\ell,k} = \sqrt{2}(-1)^k \, \Im(\psi_{\ell,|k|})$ if $k<0$ and $\varphi_{\ell,k} = \sqrt{2}(-1)^k \, \Re(\psi_{\ell,k})$ if $k>0$.

\subsection{Number of Spherical Harmonics}
\label{sec:cv_number}

\begin{figure}[h!]
  \centering
  \subfigure[$N_{\ell}^d$]{
  \includegraphics[scale=0.46]{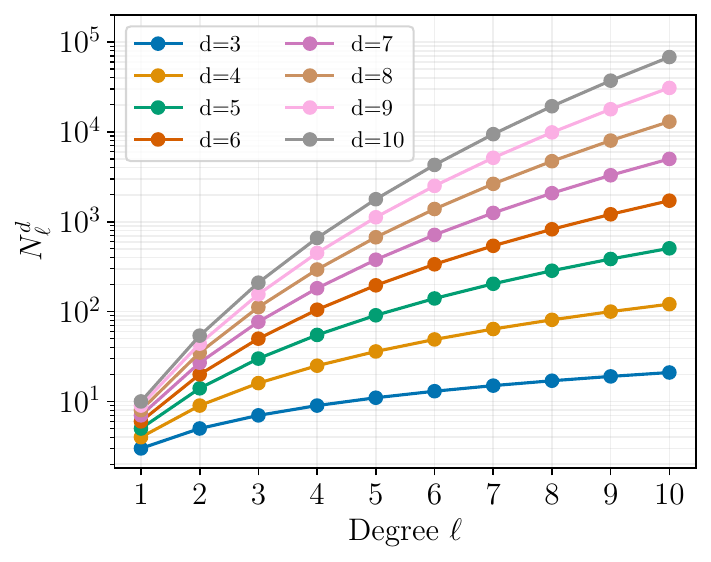}\label{fig:nb_harmo}}
   \subfigure[$\sum_{\ell=1}^L N_{\ell}^d$]{
  \includegraphics[scale=0.46]{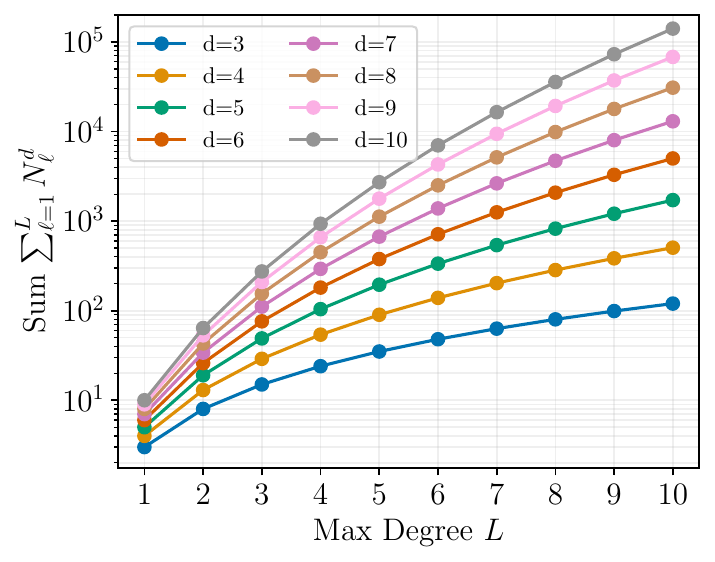}\label{fig:cumsum_harmo}}
   \subfigure[$\sum_{\ell=1}^L N_{2\ell}^d$]{
  \includegraphics[scale=0.46]{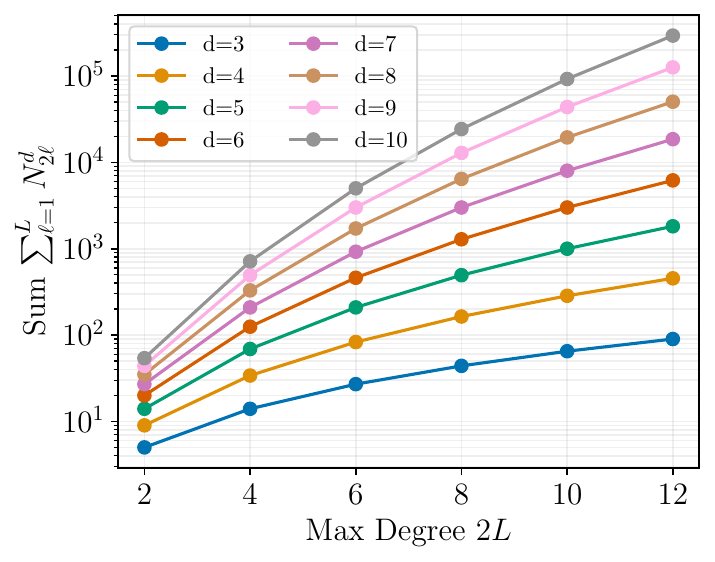}\label{fig:even_cumsum_harmo}}
\vspace{-0.7cm}
\caption{Evolution of the number $N_{\ell}^d$ of Spherical Harmonics and (even) cumulative sums according to the degree.}
    \label{fig:nb_spherical}
\end{figure}

\cref{fig:nb_harmo} presents the evolution of the number $N_{\ell}^d$ of spherical harmonics of degree $\ell$ in dimension $d$ with $d \in [3,10]$ and $\ell \in [1,10]$ while \cref{fig:cumsum_harmo,fig:even_cumsum_harmo} report the evolution of the cumulative sum $\sum_{\ell=1}^L N_{\ell}^d$ and even cumulative sum $\sum_{\ell=1}^L N_{2\ell}^d$ for maximum degree $L \in [1,10]$ and $2L \in \{2;4;6;8;10;12\}$ respectively.

\Cref{tab:cv_number} reports the maximal degree of spherical harmonics and number of control variates for the considered dimensions in the experiments (See \Cref{lem:counting}). For $d \in \{10;20\}$ we only consider a subset of spherical harmonics of degree $4$ to obtain a total number of $s=300$ control variates.
\begin{table}[h]
    \centering
    \begin{tabular}{lccccc}
    \toprule
       dimension $d$ &  $3$ & $5$ & $6$ & $10$ & $20$ \\
       \midrule
       max degree $2L$ & $16$ & $6$ & $4$ & $4$ & $4$ \\
       number $s$ of control variates & $152$  & $209$ & $125$  & $714$ & $8854$ \\
       \bottomrule
    \end{tabular}
    \caption{Parameter setting of control variates.}
    \label{tab:cv_number}
\end{table}

\subsection{Spherical Harmonics and Gegenbauer Polynomials}

For $\alpha>0$, Gegenbauer polynomials of degree $\ell$, denoted by $C_{\ell}^{\alpha}:[-1,1] \rightarrow \mathbb{R}$ are orthogonal polynomials with respect to the weight function $z \mapsto \left(1-z^2\right)^{\alpha-1 / 2}$. They are special cases of Jacobi polynomials and can be written as
$$
C_{\ell}^{\alpha}(z)=\sum_{k=0}^{\lfloor\ell / 2\rfloor} \frac{(-1)^k \Gamma(\ell-k+\alpha)}{\Gamma(\alpha) k! (\ell-2 k)!}(2 z)^{\ell-2 k}, \qquad C_{\ell}^{\alpha}(1) = \frac{\Gamma(2\alpha + \ell)}{\Gamma(2\alpha)\ell!}\cdot
$$
In the following we fix $\ell$, set $N=N_{\ell}^d$, and fix $\left\{\varphi_1, \ldots, \varphi_N\right\}$ as an orthonormal basis for $\mathscr{H}_{\ell}^d$. Denote by $Z_{\ell}(\cdot,\cdot)$ the reproducing kernel of $\mathscr{H}_{\ell}^d$. In terms of orthonormal basis we have 
\begin{align*}
    \forall \theta, \eta \in \sphere, \quad Z_{\ell}(\theta,\eta) = \sum_{k=1}^N \varphi_k(\theta) \varphi_k(\eta).
\end{align*}
Interestingly, the spherical harmonics and linked to the Gegenbauer polynomials through the following addition formula
\begin{theorem}\citep[Theorem 1.2.6]{dai2013approximation} For $\ell \geq 0$ and $\theta,\eta \in \sphere$, $d \geq 3$
\begin{align*}
    Z_{\ell}(\theta,\eta) = \sum_{k=1}^{N} \varphi_{k}(\theta) \varphi_{k}(\eta) = \frac{\ell + \alpha}{\alpha} C_{\ell}^\alpha(\langle \theta, \eta \rangle), \qquad \alpha = \frac{d-2}{2}\cdot
\end{align*}
\end{theorem}

\begin{corollary}\citep[Corollary 1.2.8]{dai2013approximation}
For $\alpha=\frac{d-2}{2}$, the Gegenbauer polynomials $C_\ell^{\alpha}$ satisfy the orthogonality relation
$$
\frac{|\mathbb{S}^{d-2}|}{|\sphere|} \int_{-1}^1 C_\ell^\alpha(t) C_{\ell'}^\alpha(t)\left(1-t^2\right)^{\alpha-\frac{1}{2}} \diff t=h_{\ell}^\alpha \delta_{\ell,\ell'}, \qquad
h_{\ell}^\alpha=\frac{\alpha}{\ell+\alpha} C_{\ell}^\alpha(1).$$
\end{corollary}

Let $\left\{\theta_1, \ldots, \theta_N\right\}$ be a collection of points on $\sphere$. The following remarks motivate the definition of a fundamental set. We let $M_1:=\varphi_1\left(\theta_1\right)$ and for $k=2,3, \ldots, N$, define matrices
$$
M_k:=\left[\begin{array}{ccc}
\varphi_1\left(\theta_1\right) & \ldots & \varphi_1\left(\theta_k\right) \\
\vdots & \ldots & \vdots \\
\varphi_k\left(\theta_1\right) & \ldots & \varphi_k\left(\theta_k\right)
\end{array}\right], \quad M_k(\theta):=\left[\begin{array}{c|c} 
& \varphi_1(\theta) \\
M_{k-1} & \vdots \\
& \varphi_{k-1}(\theta) \\
\hline \varphi_k\left(\theta_1\right), \ldots \varphi_k\left(\theta_{k-1}\right) & \varphi_k(\theta),
\end{array}\right]
$$
The product of $M_N$ and its transpose $M_N^\top$ can be summed, on applying the addition formula, as $M_N^\top M_N=\left[Z_{\ell}\left(\theta_i, \theta_j\right)\right]_{i, j=1}^N$, which shows, in particular, $\operatorname{det}\left[Z_{\ell}\left(\theta_i, \theta_j\right)\right]_{i, j=1}^N=\left(\operatorname{det} M_N\right)^2 \geq 0
$

\begin{definition}\citep[Definition 1.3.1]{dai2013approximation}
 A collection of points $\left\{\theta_1, \ldots, \theta_N\right\}$ in $\mathbb{S}^{d-1}$ is called a fundamental system of degree $\ell$ on the sphere $\mathbb{S}^{d-1}$ if
$$
\operatorname{det}\left[C_\ell^\alpha\left(\left\langle \theta_i, \theta_j\right\rangle\right)\right]_{i, j=1}^N>0, \quad \alpha=\frac{d-2}{2}
$$    
\end{definition}

\begin{theorem}\citep[Theorem 1.3.3]{dai2013approximation}
    If $\left\{\theta_1, \ldots, \theta_N\right\}$ is a fundamental system of points on the sphere, then $\{C_{\ell}^{\alpha}(\langle \cdot, \theta_i \rangle)| i=1,\ldots,N\}$, $\alpha=(d-2)/2$ is a basis of $\mathscr{H}_{\ell}^d|_{\sphere}$
\end{theorem}

This theorem lies at the heart of the practical implementation  used in \citet{dutordoir2020sparse} which constructs a fundamental system of vectors for efficient evaluation of spherical harmonics. The mathematical aspects of this process involve concepts such as the Gegenbauer polynomial, Cholesky decomposition, optimization, and normalization. The goal is to construct a set of vectors efficiently spanning the space required for evaluating spherical harmonics in the given dimension and degree. The procedure works as follows : Start by defining the Gegenbauer polynomial, denoted as \(C_{\ell}^\alpha(\theta)\), of a specific degree $\ell$ and a dimension-dependent parameter $\alpha=(d-2)/2$. Select the first vector arbitrarily, often chosen as the north pole or the unit vector along the last dimension. Compute the Cholesky decomposition of the Gegenbauer-gram matrix associated with the current vectors and iteratively add new vectors to to expand the fundamental system. For each new vector, optimize the determinant of the Gegenbauer-gram matrix, measuring the span of the space generated by the vectors. The optimization process involves finding a new vector that maximizes the determinant of the augmented matrix. This is typically done using optimization techniques such as BFGS. After finding the optimal vector, normalize it to ensure it has unit length.

\subsection{Simple Case Study for Discrete Measures and Gaussian Measures} \label{subsec:case_study}

To investigate the approximation power of the integrand $\theta \mapsto \fmn(\theta) = \Wass_p^p(\theta_{\sharp}^\star \mu_2,\theta_{\sharp}^\star \nu_2)$ via spherical harmonics, we consider a case study in dimension $d=2$ on the unit circle $\mathbb{S}^1$ with circular harmonics $(\varphi_{\ell,k})$ (see \Cref{subsec:spherical_small_dim}) . Using an Ordinary Least Squares regression, the polynomial approximation $\widehat f_{2L}$ of $\fmn$ can be written as 
\begin{equation*}
    \widehat{f}_{2L}(\theta) = \alpha + \sum_{\ell = 2, 4, \ldots, 2L} \left(\beta_{\ell}^{(1)} \varphi_{\ell,1}(\theta) + \beta_{\ell}^{(2)} \varphi_{\ell,2}(\theta)\right)
\end{equation*}
where $\alpha$ and $\beta = (\beta_2^{(1)},\beta_2^{(2)},\ldots)$ are estimated by the OLS procedure. We investigate the smoothness properties of the integrand $\fmn$ by comparing different measures $\mu$ and $\nu$. We first consider the case of discrete measures $\mu_2$ and $\nu_2$ each supported on two atoms. In this simple case, the integrand $\fmn(\theta)$ is only Lipschitz and we study the approximation error of our method. Then we consider the case of Gaussian measures $\mu = \mathcal{N}_2(\mathrm{m}_X,\mathbf{\Sigma}_X)$ and $\nu = \mathcal{N}_2(\mathrm{m}_Y,\mathbf{\Sigma}_Y)$. In that case, the integrand of interest is infinitely smooth.

\textbf{Discrete measures.} We consider the discrete measures $\mu_2$ and $\nu_2$ associated to the points $x_1=(0,0)$, $x_2=(1,0)$ and $y_1=(0,0)$, $y_2=(1,1)$ respectively. We have $\mu_2 = \frac{1}{2} \delta_{(0,0)} + \frac{1}{2}\delta_{(1,0)}$ and $\nu_2 = \frac{1}{2} \delta_{(0,0)} + \frac{1}{2} \delta_{(1,1)}$. The integrand $\fmn$ can be computed exactly as
	\[ \Wass_p^p(\theta_{\sharp}^\star \mu_2,\theta_{\sharp}^\star \nu_2) = 
	\begin{dcases} 
		\tfrac{1}{2} |\sin(\theta)|^p, & \theta \in [0,\pi/2]\cup[3\pi/4,3\pi/2]\cup[7\pi/4,2\pi], \\
		\tfrac{1}{2}( |\cos(\theta)|^p + |\cos(\theta)+\sin(\theta)|^p), & \theta \in [\pi/2,3\pi/4]\cup[3\pi/2,7\pi/4]. 
	\end{dcases}
	\]
 By focusing on the case of order $p=2$, we have
	\begin{align*}
		I_1 &= \int_{0}^{\pi/2} \sin^2(\theta) \diff \theta = \frac{\pi}{4}, &
		I_2 &= \int_{\pi/2}^{3\pi/4} \cos^2(\theta) + (\cos(\theta)+\sin(\theta))^2 \diff \theta = \frac{3 \pi}{8} - \frac{3}{4}, \\
		I_3 &= \int_{3 \pi/4}^{3 \pi/2} \sin^2(\theta) \diff \theta = \frac{3 \pi}{8} - \frac{1}{4}, &
		I_4 &= \int_{3 \pi/2}^{7\pi/4} \cos^2(\theta) + (\cos(\theta)+\sin(\theta))^2 \diff \theta = \frac{3 \pi}{8} - \frac{3}{4}, \\
		I_5 &= \int_{7 \pi/4}^{2 \pi} \sin^2(\theta) \diff \theta = \frac{ \pi}{8} - \frac{1}{4}. 
	\end{align*}
	Therefore,
	\begin{align*}
		\slice_2^2 (\mu_2, \nu_2) = \frac{1}{2 \pi} \int_{0}^{2 \pi} W_2^2(\theta_{\sharp}^\star \mu_2,\theta_{\sharp}^\star \nu_2) \diff \theta 
		= \frac{1}{4 \pi} (I_1 + I_2 + I_3 + I_4 + I_5) 
  =   \frac{3}{8} - \frac{1}{2\pi}.
	\end{align*}
	Recall that $\SHCV_{n,L}^p(\mu_2,\nu_2) = \Int_n^{\mathrm{ols}}$ where $(\Int_n^{\mathrm{ols}},\beta_n) \in \arg\min_{\alpha,\beta} \|f_n - \alpha \1_n - \Phi \beta \|_2^2$.
	
\begin{figure}[h!]
  \centering
  \subfigure[$\Wass_2^2(\theta_{\sharp}^\star \mu_2,\theta_{\sharp}^\star \nu_2)$ and $\widehat{f}_{2L}(\theta)$, discrete case]{
  \includegraphics[scale=0.53]{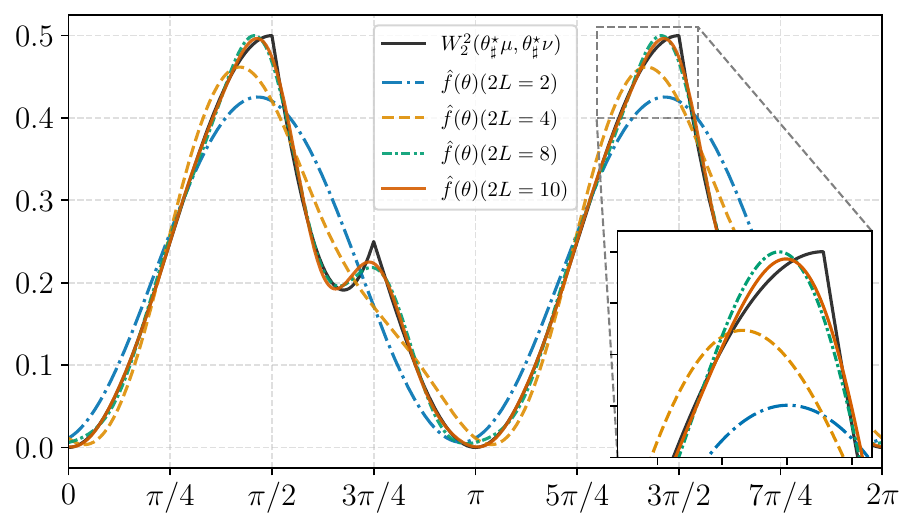}\label{fig:W2_approx_d}}
   \subfigure[Error $\varepsilon=\Wass_2^2(\theta_{\sharp}^\star \mu_2,\theta_{\sharp}^\star \nu_2)-\widehat{f}_{2L}(\theta)$]{
  \includegraphics[scale=0.53]{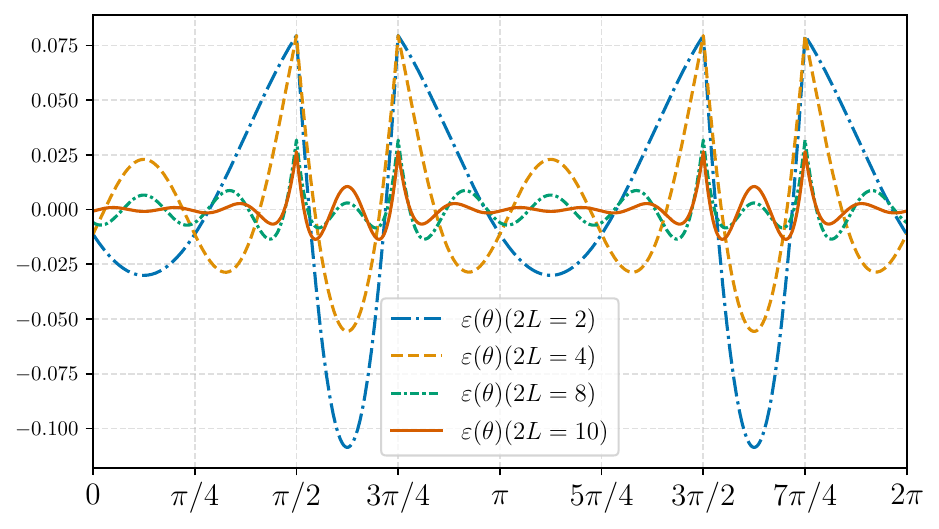}\label{fig:error_approx_d}}
  \vspace{-0.3cm}
\caption{$\Wass_2^2(\theta_{\sharp}^\star \mu_2,\theta_{\sharp}^\star \nu_2)$, polynomial approximation $\widehat{f}_{2L}(\theta)$ and errors $\varepsilon=\Wass_2^2(\theta_{\sharp}^\star \mu_2,\theta_{\sharp}^\star \nu_2)-\widehat{f}_{2L}(\theta)$ for two discrete measures.}
    \label{fig:pol_approx_w22}
\end{figure}

Figure~\ref{fig:W2_approx_d} reports the evolution of $\Wass_2^2(\theta_{\sharp}^\star \mu_2,\theta_{\sharp}^\star \nu_2)$ (black line) and its polynomial approximation $\widehat{f}_{2L}(\theta)$ with different degrees of harmonics $2L \in \{2;4;8;10\}$. Figure~\ref{fig:error_approx_d} displays the approximation error, i.e., the difference $\varepsilon_n$ between $\Wass_2^2(\theta_{\sharp}^\star \mu_2,\theta_{\sharp}^\star \nu_2)$ and its polynomial approximation $\widehat{f}_{2L}(\theta)$.

\textbf{Gaussian measures.} Let $\mu = \mathcal{N}_d(\mathrm{m}_X,\mathbf{\Sigma}_X)$ and $\nu = \mathcal{N}_d(\mathrm{m}_Y,\mathbf{\Sigma}_Y)$. Then, for $\theta_{\sharp}^\star \mu = \mathcal{N}_1(\theta^\top \mathrm{m}_X, \theta^\top \mathbf{\Sigma}_X \theta)$ and $\theta_{\sharp}^\star \nu = \mathcal{N}_1(\theta^\top \mathrm{m}_Y, \theta^\top \mathbf{\Sigma}_Y \theta)$, we have $\Wass_2^2(\theta_{\sharp}^\star \mu,\theta_{\sharp}^\star \nu) = \left| \theta^\top (\mathrm{m}_X - \mathrm{m}_Y)\right|^2 + (\sqrt{\theta^\top \mathbf{\Sigma}_X \theta} -\sqrt{\theta^\top \mathbf{\Sigma}_Y \theta} )^2$.
We consider $\mathrm{m}_X = (0,0)^\top$, $\mathrm{m}_Y=(1,1)^\top$, $\Sigma_X = 
\begin{bmatrix}
	1 & 0.2  \\
	0.2 & 1 
\end{bmatrix}$, $\Sigma_Y = 
\begin{bmatrix}
	10 & 3  \\
	3 & 10 
\end{bmatrix}$

\begin{figure}[h!]
  \centering
  \subfigure[$\Wass_2^2(\theta_{\sharp}^\star \mu_2,\theta_{\sharp}^\star \nu_2)$ and $\widehat{f}_{2L}(\theta)$, gaussian case]{
  \includegraphics[scale=0.55]{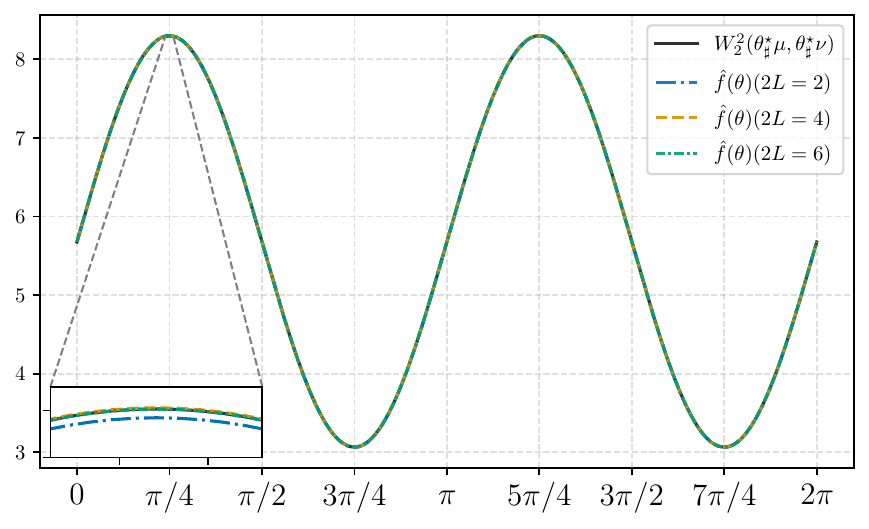}\label{fig:gaussian_approx}}
   \subfigure[Error $\varepsilon=\Wass_2^2(\theta_{\sharp}^\star \mu_2,\theta_{\sharp}^\star \nu_2)-\widehat{f}_{2L}(\theta)$]{
  \includegraphics[scale=0.55]{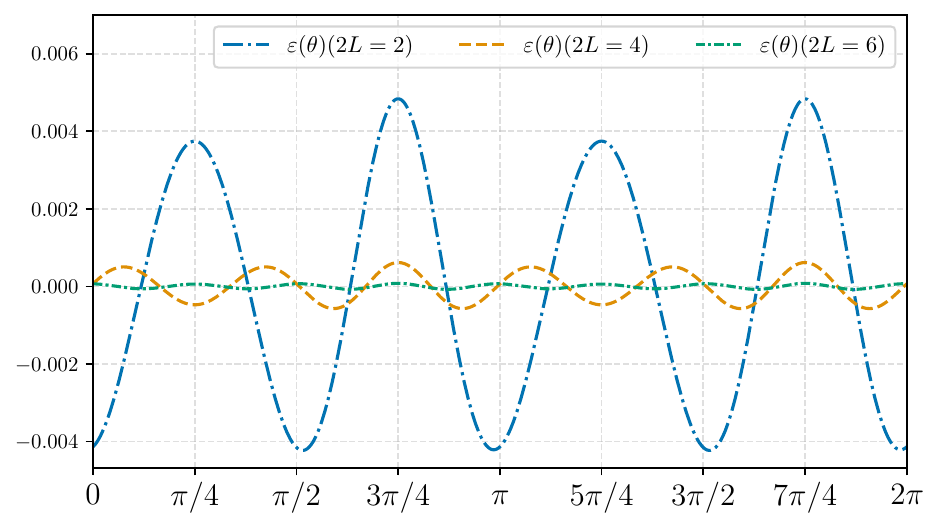}\label{fig:error_approx_gaussian}}
  \vspace{-0.3cm}
\caption{$\Wass_2^2(\theta_{\sharp}^\star \mu_2,\theta_{\sharp}^\star \nu_2)$, polynomial approximation $\widehat{f}_{2L}(\theta)$ and errors $\varepsilon=\Wass_2^2(\theta_{\sharp}^\star \mu_2,\theta_{\sharp}^\star \nu_2)-\widehat{f}_{2L}(\theta)$ for two Gaussian measures.}
    \label{fig:gauss_approx_w22}
\end{figure}

Figure \ref{fig:gaussian_approx} reports the evolution of $\Wass_2^2(\theta_{\sharp}^\star \mu_2,\theta_{\sharp}^\star \nu_2)$ (black line) and its polynomial approximation $\widehat{f}_{2L}(\theta)$ with different degrees of harmonics $2L \in \{2;4;6\}$. Observe that in this case the integrand is very smooth so that the polynomial approximations seem to overlap. Figure \ref{fig:error_approx_gaussian} displays the approximation error, \textit{i.e.}, the difference $\varepsilon_n$ between $\Wass_2^2(\theta_{\sharp}^\star \mu_2,\theta_{\sharp}^\star \nu_2)$ and its polynomial approximation $\widehat{f}_{2L}(\theta)$.

\begin{figure}[h!]
  \centering
   \subfigure[Boxplots Gaussian vs Discrete]{
  \includegraphics[scale=0.6]{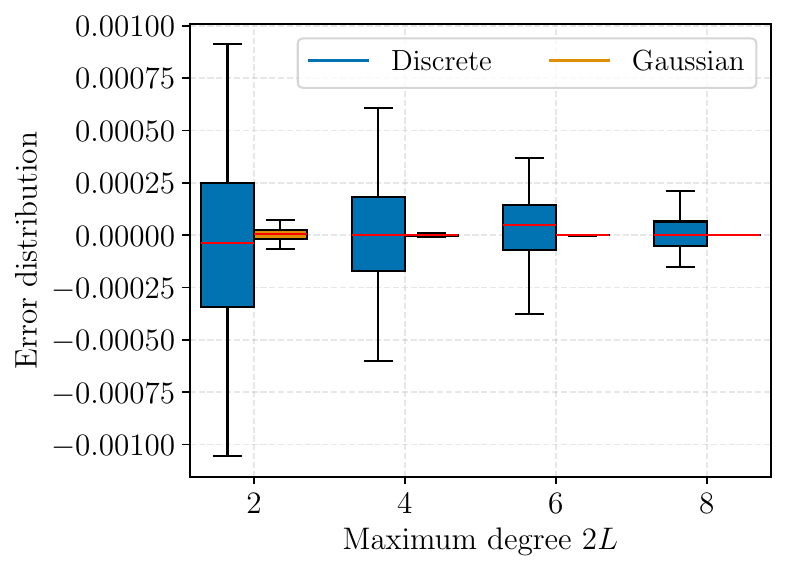}\label{fig:box_compare}}
  \hspace{1cm}
  \subfigure[Boxplot Gaussian]{
  \includegraphics[scale=0.6]{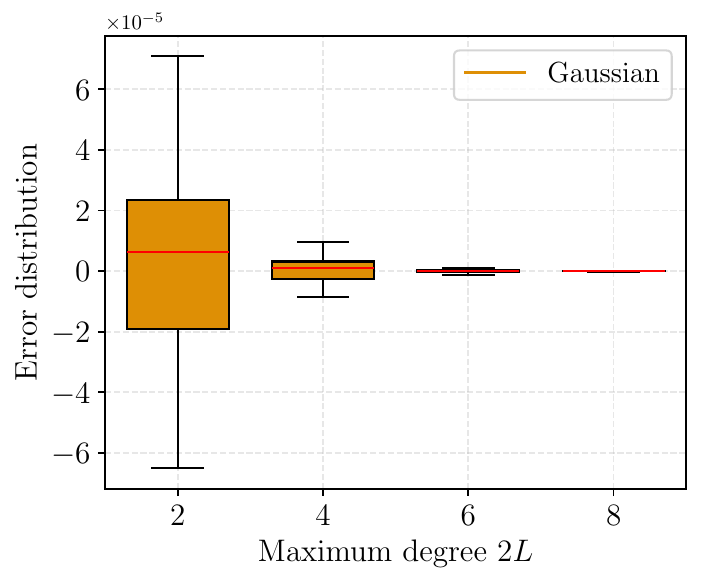}\label{fig:box_gauss}}
\caption{Error distribution of $(\mathrm{SHCV}_{n,L}^2-\slice_2^2)$.}
    \label{fig:boxplots_approximation}
\end{figure}

Figure \ref{fig:box_compare} presents the boxplots of the error distribution $(\mathrm{SHCV}_{n,L}^2-\slice_2^2)$ both for discrete measures $\mu_2, \nu_2$ and for Gaussian measures   with different degrees of harmonics $2L \in \{2;4;6;8\}$ and Monte Carlo sample size $n=10^4$. The Sliced-Wasserstein distance for Gaussian measures is obtained by the naive Monte Carlo method with $n=10^8$. The boxplots are obtained over $100$ replications.

\clearpage
\subsection{Numerical Check for Exact Integration Rule} \label{subsec:check}

\begin{figure}[h!]
  \centering
  \subfigure[$d=3$]{
  \includegraphics[scale=0.62]{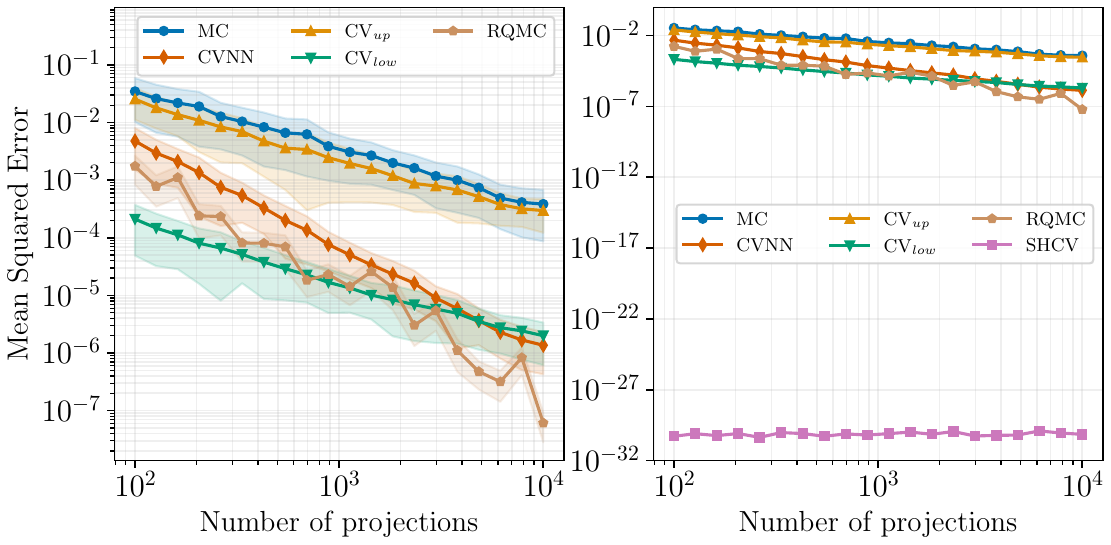}\label{fig:check_d3}}
   \subfigure[$d=5$]{
  \includegraphics[scale=0.62]{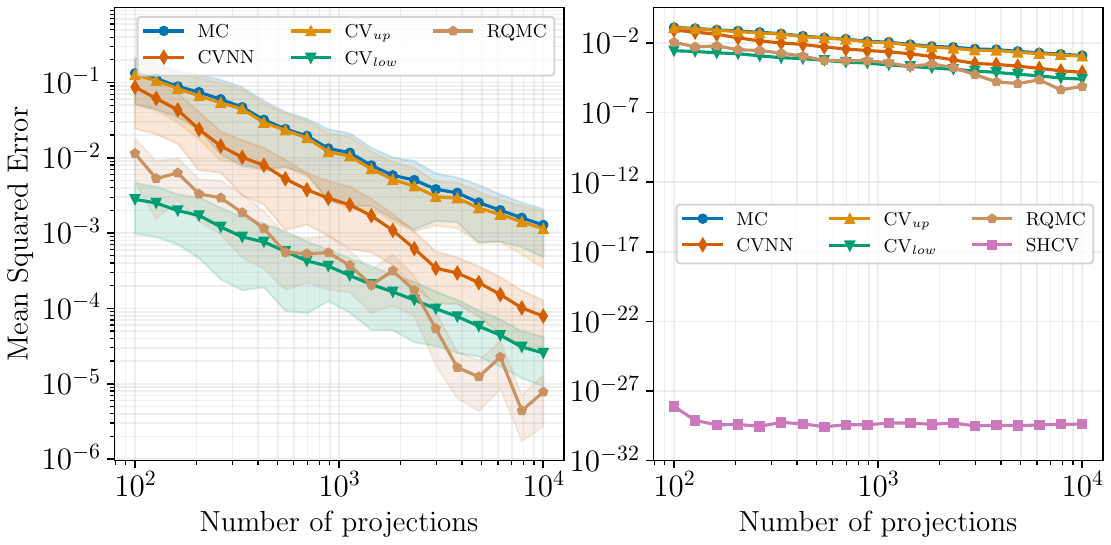}\label{fig:check_d6}}
\caption{MSE for Gaussian distributions with proportional dispersion matrices, dimension $d\in \{3;5\}$, obtained over $100$ replications.}
    \label{fig:check_numerical}
\end{figure}

We compute $\slice_2^2(\mu,\nu)$ for $\mu=\Normal(a,\mathbf{A})$ and $\nu=\Normal(b,\mathbf{B})$ with means $a,b \sim \Normal(\1_d,I_d)$ and proportional (see \Cref{subsec:elem}) covariance matrices $\mathbf{A}=\Sigma_a^{}\Sigma_a^\top$ and $\bB=\gamma \bA$ where all the entries of $\Sigma_a$ are drawn according to $\Normal(0,1)$. We consider dimensions $d\in \{3;5\}$ and a number of projections $n \in [10^2;10^4]$. Observe that in this case, the integrand of interest is a quadratic polynomial in $\theta$ given by
\begin{align*}
f_{\mu,\nu}^{(2)}(\theta)  
&= |\theta^\top(a-b)|^2 + \bigl(\sqrt{\theta^\top \bA \theta}- \sqrt{\theta^\top \bB \theta}\bigr)^2 \\
&= |\theta^\top(a-b)|^2 + (1-\sqrt{\gamma})^2 (\theta^\top \bA \theta),
\end{align*}
and $\slice_2^2(\mu,\nu)$ is given in close-form by integrating over $\theta \in \sphere$ such that
\begin{align*}
    \slice_2^2(\mu,\nu) = \frac{\|a-b\|_2^2}{d} + \frac{(1-\sqrt{\gamma})^2}{d}\mathrm{Tr}(\bA).
\end{align*}
\Cref{fig:check_numerical} reports the mean squared errors for the different estimates $\widehat{\slice}_{n}(\mu,\nu)$ where the expectation is computed as an average over $100$ independent runs with parameter $\gamma=2$. Observe that, while the standard Monte Carlo estimate and other control variate-based methods achieve a small MSE in the range $[10^{-7};10^{-3}]$, our SHCV estimate obtains a MSE of $10^{-30}$ which is virtually zero up to machine precision. This numerical comparison empirically validates the exact integration rule of \Cref{prop:exact}.

\end{document}